\documentclass{article}

\usepackage{arxiv}

\usepackage[utf8]{inputenc} %
\usepackage[T1]{fontenc}    %
\usepackage{hyperref}       %
\usepackage{url}            %
\usepackage{enumitem}
\usepackage{booktabs}       %
\usepackage{amsfonts}       %
\usepackage{nicefrac}       %
\usepackage{microtype}      %
\usepackage{mathtools}
\usepackage{wrapfig}
\usepackage{bm}
\usepackage{lipsum}		%
\usepackage{graphicx}
\usepackage{natbib}
\usepackage{adjustbox}

\usepackage{tikz}
\tikzstyle{arrow} = [thick,->,>=stealth,line width=0.5pt]

\usepackage{algorithm}
\usepackage{algorithmic}
\usepackage{subcaption}
\usepackage{amsthm}
\usepackage{amssymb}
\usepackage{multirow}

\usepackage{ifthen}

\title{Treatment Learning Causal Transformer for Noisy Image Classification}

\author{ Chao-Han Huck Yang$^*$\\
Georgia Institute of Technology\\
huckiyang@gatech.edu
\And
I-Te Danny Hung$^*$\\
Columbia University\\
ih2320@columbia.edu
\And
Yi-Chieh Liu$^*$\\
Georgia Institute of Technology\\
yliu3233@gatech.edu\\
\And
Pin-Yu Chen\\
IBM Research\\
pin-yu.chen@ibm.com
}

\begin{document}

\maketitle

\begin{abstract}%
Current top-notch deep learning (DL) based vision models are primarily based on exploring and exploiting the inherent correlations between training data samples and their associated labels. However, a known practical challenge is their degraded performance against ``noisy'' data, induced by different circumstances such as spurious correlations, irrelevant contexts, domain shift, and adversarial attacks. In this work, we incorporate this binary information of ``existence of noise'' as treatment into image classification tasks to improve prediction accuracy by jointly estimating their treatment effects. Motivated from causal variational inference, we propose a transformer-based architecture, Treatment Learning Causal Transformer (TLT), that uses a latent generative model to estimate robust feature representations from current observational input for noise image classification. Depending on the estimated noise level (modeled as a binary treatment factor), TLT assigns the corresponding inference network trained by the designed causal loss for prediction. We also create new noisy image datasets incorporating a wide range of noise factors (e.g., object masking, style transfer, and adversarial perturbation) for performance benchmarking. The superior performance of TLT in noisy image classification is further validated by several refutation evaluation metrics. As a by-product, TLT also improves visual salience methods for perceiving noisy images.

\end{abstract}

\section{Introduction}

Although deep neural networks (DNNs) have surpassed human-level ``accuracy'' in many image recognition tasks~\citep{ronneberger2015u,he2016deep,huang2017densely}, current DNNs still implicitly rely on the \emph{assumption}~\citep{pearl2019seven} on the existence of a strong correlation between training and testing data. Moreover, increasing evidence and concerns~\citep{alcorn2019strike} show that using the correlation association for prediction can be problematic against \textit{noisy} images~\citep{xiao2015learning}, such as pose-shifting of identical objects~\citep{alcorn2019strike} or imperceptible perturbation~\citep{goodfellow2014explaining}. 
In practice, real-world image classification often involves rich, noisy, and even chaotic contexts, intensifying the demand for generalization in the wild. Putting in a unified descriptive framework,

\begin{figure}[t]
\centering

\begin{subfigure}{0.15\textwidth} 
\centering
		\includegraphics[width=\textwidth]{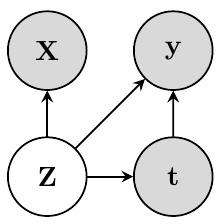}
\label{fig:figure1c}
\end{subfigure}
\quad
\begin{subfigure}{0.50\textwidth} %
	    \centering
		\includegraphics[width=\textwidth]{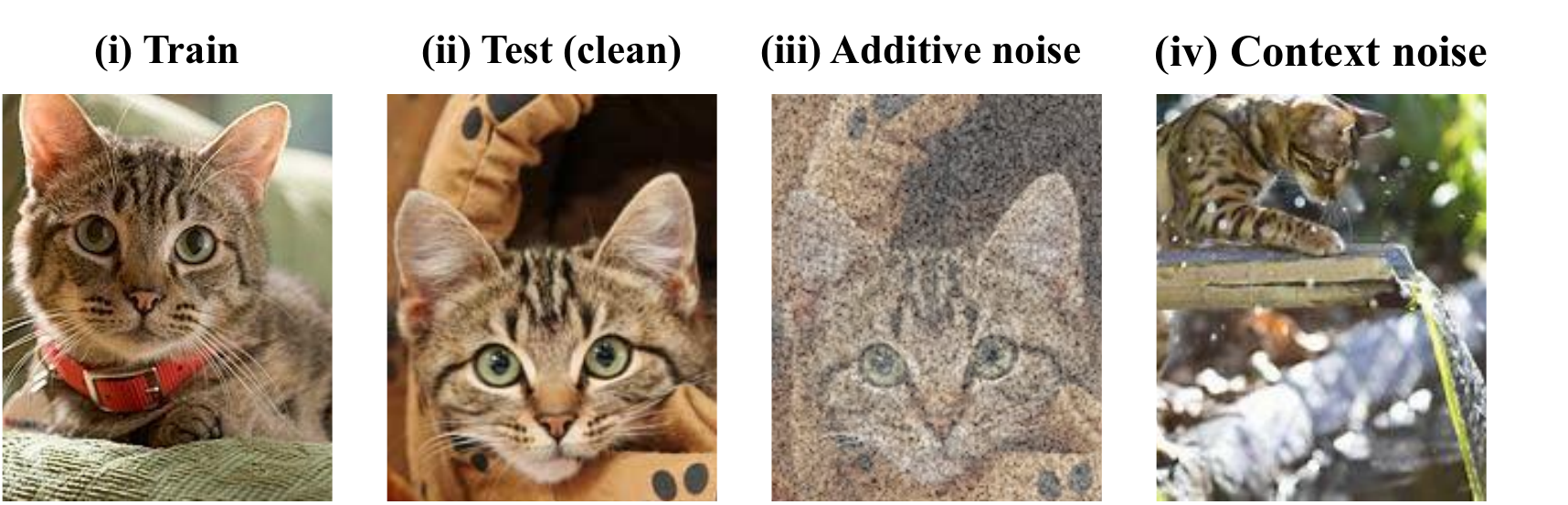}
		\label{fig:figure1a}
	\end{subfigure}
\caption{ \emph{(a)} An example of deployed causal graphical model (CGM), where $\mathbf{Z}$ denotes unobservable confounder variable (e.g., the concept of ``cat''), $\mathbf{X}$ denotes a noisy observation of confounder (e.g., an image can still be recognized as a cat), $\mathbf{y}$ denotes outcome (e.g., a label), and $\mathbf{t}$ denotes the information of a binary treatment (e.g., the existence of extra semantic patterns or additive noise; thus, it is equal to $0$ or $1$), which is \textbf{observable} during \textbf{training} and \textbf{unobservable} during \textbf{testing} time. \emph{(b)} Images with ``cat'' labels, where (i) and (ii) share the same context of ``indoor''; (iii) shows a noisy setup of (ii) undergoing additive Gaussian perturbation; (iv) shows another setup of introducing extra noisy semantic patterns (e.g., ``waterside'') in NICO~\citep{he2020towards} noisy images dataset.} 
\label{fig:figure1}
\vspace{-4mm}
\end{figure}

To address machine perception against noisy images, we are inspired by how human performs visual recognition. Human's learning processes are often mixed with logic inference (e.g., a symbolic definition from books) and representation learning (e.g., an experience of viewing a visual pattern). One prominent difference between current DNNs and human recognition systems is the capability in causal inference. Mathematically,
causal learning \citep{pearl1995causal, peters2014causal} is a statistical inference model that infers beliefs or probabilities under uncertain conditions, which aims to identify latent variables (called ``confounders'') that influence both intervention and outcome. The unobserved confounders may be abstract in a cognitive-level (e.g., concepts) but could be observed via their noisy view in the real-world (e.g., objects). For instance, as shown in Fig.~\ref{fig:figure1} \emph{(a)}, confounder learning aims to model a prediction process by finding a representation (e.g., ``cat'') and avoiding relying on irrelevant patterns (e.g., ``waterside''). Intuitively, with causal modeling and confounder inference, correct prediction can be made on noisy inputs, where the generative estimation process, such as causal effect variational autoencoder (CEVAE)~\citep{louizos2017causal}, affects multiple covariates for predicting data profiles. In this work, we aim to incorporate the effects of causal confounder learning to image classification, as motivated by cognitive psychology for causal learning. Specifically, we use the attention mechanism for noise-resilience inference from patterns. We design a novel sequence-to-sequence learning model, \textbf{Treatment Learning Causal Transformer (TLT)}, which leverages upon the conditional query-based attention and the inference power from a variational causal inference model.

Our TLT tackles noisy image classification by jointly learning to a generative model of $\mathbf{Z}$ and estimating the effects from the treatment information ($\mathbf{t}$), as illustrated in Fig.~\ref{fig:figure1} (a). This model consists of unobservable confounder variables $\mathbf{Z}$ corresponding to the ground-truth but inaccessible information (e.g., the ontological concept~\citep{trampusch2016between} of a label), input data $\mathbf{X}$ from a noisy view of $\mathbf{Z}$ (e.g., images), a treatment~\citep{pearl2016causal} information  $\mathbf{t}$ given $\mathbf{X}$ and $\mathbf{Z}$ (e.g., secondary information as visual patterns and additive noise without directly affecting our understanding the concept of ``cat''), and a classification label $\mathbf{y}$ from the unobservable confounder. 
Built upon this causal graphical model, our contributions are:
\vspace{-2mm}
\begin{itemize}[leftmargin=*]
    \item A transformer architecture (TLT) for noisy image classification are presented, which is based on a treatment estimation architecture and a causal variational generative model with competitive classification performance against noisy image. 
    \item We further curated a new noisy images datasets, Causal Pairs (CPS), to study generalization under different artificial noise settings for general and medical images.
    \item We use formal statistical refutations tests to validate the causal effect of TLT, and show that TLT can improve visual saliency methods on noisy images.
\end{itemize}

\section{Related Work} 
\textbf{Noisy Image Classification.}
Prior works on noisy images classification have highlighted the importance of using generative models \citep{oquab2014learning} to ameliorate the negative learning effects from noisy data. Xiao \emph{et al.}~\citep{xiao2015learning} leverage a conditional generative model~\citep{sohn2015learning} to capture the relations among images and noise types from online shopping systems. Direct learning from noisy data is another approach by using statistical sampling~\citep{han2019deep, li2017learning} and active learning~\citep{gal2017deep} for performance enhancement. Meanwhile, new noisy images dataset and evaluation metrics~\citep{he2020towards} on context independence have been proposed, such as Strike simulator \citep{alcorn2019strike} for synthesizing pose-shifting images and NICO~\citep{he2020towards, zhang2021deep, liu2021stable} as the open-access noisy image dataset. NICO further aims to highlight the importance of incorporating a statistical inference (e.g., causal model) for improved image classification with large-scale noisy context-patterns (e.g., an image shows ``cat in waterside'' but given a single label of ``cat''). However, different from context-wise noise in NICO, modeling sizeable artificial noise in images is crucial yet remains unexplored. In this work, we create a new image dataset containing various artificial noise and use the NICO~\citep{he2020towards} with a generative causal model for performance benchmarking. \\
\textbf{Causal Learning for Computer Vision.} 
Many efforts \citep{pickup2014seeing, fire2016learning, lebeda2015exploring, fire2013using} have leveraged upon causal learning to better understand and interpret toward vision recognition tasks. Lopez-Paz \emph{et al.} \citep{lopez2017discovering} propose utilizing DNNs to discover the causation between image class labels for addressing the importance of this direct causal relationship affecting model performance and context grounding. Incorporating causal analysis and regularization showed improved performance in generative adversarial models such as Causal-GANs~\citep{kocaoglu2017causalgan, bahadori2017causal}. 
However, infusing causal modeling and inference to DNN-based image recognition systems is still an open challenge. For instance, in previous works \citep{lopez2017discovering, yang2019causal}, researchers focus on modeling a direct causal model (DCM)~\citep{pearl2016causal} for visual learning. The DCMs treat a visual pattern (e.g., texture) as a cause visual representation (e.g., patterns of the ``cat'') and barely incorporate additional label information (e.g., context) or apply noise as a \textbf{treatment} in causal analysis. In recent works, causal modeling also show promising results in a large-scale computer vision task, such scene graph~\citep{tang2020unbiased} generation, visual and language learning~\citep{qi2020two,agarwal2020towards,abbasnejad2020counterfactual}, and semantic segmentation~\citep{zhang2020causal}. The work of Chalupkaet al. \citep{chalupka2014visual} is closer to our work by deploying interventional experiments to target causal relationships in the labeling process. However, modeling the aforementioned treatment effects and designing efficient learning models are still not fully explored~\citep{pearl2019seven}.\\
\textbf{Causal Inference by Autoencoder.} 
Recently, classical causal inference tasks, such as regression modeling~\citep{buhlmann2014cam}, risk estimation~\citep{pearl2019seven}, and causal discovery~\citep{monti2020causal}, have been incorporated with deep generative models~\citep{rezende2014stochastic} and attained state-of-the-art performance~\citep{shalit2017estimating, louizos2017causal}. These generative models often use an encoder-decoder architecture to improve both logic inference and features extracted from a large-scale dataset with noisy observations. TARNet~\citep{shalit2017estimating} is one foundational DNN model incorporating causal inference loss from a causal graphical model (CGM) and feature reconstruction loss jointly for linear regression, showing better results compared with variational inference models~\citep{kingma2013auto}. 
Inspired by the CGM of TARNet~\citep{shalit2017estimating}, causal-effect variational autoencoder~(CEVAE) was proposed in \citep{louizos2017causal, yang2021causal} for regression tasks, which draws a connection between causal inference with proxy variables and latent space learning for approximating the hidden and unobservable confounder by the potential outcome model from Rubin's causal inference framework~\citep{imbens2010rubin, rubin1974estimating}. 
Our proposed causal model in TLT shares a similar CGM with CEVAE but has a different training objective, probabilistic encoding, and specific design for visual recognition, such as the use of attention mechanism. 

\section{TLT: Treatment Learning Causal Transformer}

\begin{wraptable}{r}{0.48\textwidth}
\vspace{-4mm}
\caption{\emph{Causal hierarchy}~\citep{pearl2009causality}: questions at level $i$ can only be answered if information from the same or higher level is available.}
\label{tab:1:causal:hi}
\begin{adjustbox}{width=0.47\textwidth}
\begin{tabular}{|l|l|l|l|}
\hline
\textbf{Level} & \textbf{Activity} & \textbf{PGM} & \textbf{Example} \\ \hline
($\mathbb{I}$)  Association & Observing & $P(y|x)$ & ResNet~\citep{he2016deep} \\ \hline
($\mathbb{II}$)  Intervention & Intervening & $P(y|do(x),z)$ & TLT (ours) \\ \hline
\end{tabular}
\end{adjustbox}
\vspace{-2mm}
\end{wraptable}

\subsection{Modeling under Causal Hierarchy Theorem}
\label{sec:3:1:modeling}
To model a general image classification problem with causal inference, we introduce Pearl's \emph{causal hierarchy Theorem} \citep{bareinboim2020pearl, shpitser2008complete, pearl2009causality} as shown in Tab.~\ref{tab:1:causal:hi}, with a non-causal classification model and a causal inference model. Non-causal model is in level ($\mathbb{I}$) of causal hierarchy, which associates the $outcome$ (prediction) to the input directly by $P(y|x)$ from supervised model such as ResNet~\citep{he2016deep}. Non-causal model could be unsupervised by using approximate inference such as variational encoder-decoder~\citep{bahuleyan2018variational} with two parameterized networks, $\Theta$ and $\Phi$.
The association-level (non-causal) setup in the causal hierarchy can solve visual learning tasks at level ($\mathbb{I}$), such as non-noisy image classification. 

For noisy image classification, we argue that the problem setup is elevated to level ($\mathbb{II}$) of the causal hierarchy, requiring the capability of confounder learning and the $do$-$calculus$~\citep{pearl2019seven} (refer to causal inference foundations \textbf{supplement}~\ref{sup:causal}). We first make a formal definition on a
pair of $i^{th}$ query $(x_i,y_i)$ including a noisy image input ($x_i$) and its associated label ($y_i$). Suppose for every noisy image, there exists a clean but inaccessible image ($\tilde x_i$) and treatment information ($t_i$), where the intervened observation is modeled as $P(x_i) = P(do(\tilde x_i)) \equiv  P(\tilde x_i | t_i)$, and $t_i$ encodes full information of the intervention through the do-operator notation $do(\cdot)$. The corresponding confounder $z_i$ follows $P(z_i) = P(\tilde x_i, t_i, \tilde z_i)$, where $\tilde z_i$ is the unobservable part (e.g., undiscovered species of ``cat'' but belong to its ontological definition) of the confounder. To make a prediction ($y_i$) of a noisy input of ($x_i$), we could have the intervened view of the question by:
\begin{equation}
P(y_i|x_i)=P(y_i|do(\tilde x_i),z_i)= P(y_i|\tilde x_i, t_i, z_i)
\label{eq:c:h:do}
\end{equation}
with do-operator in level ($\mathbb{II}$) of the causal hierarchy. Based on the causal hierarchy, we could use the model with the proxy variables ($z_i, t_i$) in the higher level ($\mathbb{II}$) to answer the question in equal or lower level. Next, we introduce our training objective using an encoder-decoder architecture to reparameterize the aforementioned proxy variables for causal learning. 

\begin{figure*} %
\begin{center}
\begin{subfigure}{0.57\linewidth}
 \includegraphics[width=\linewidth]{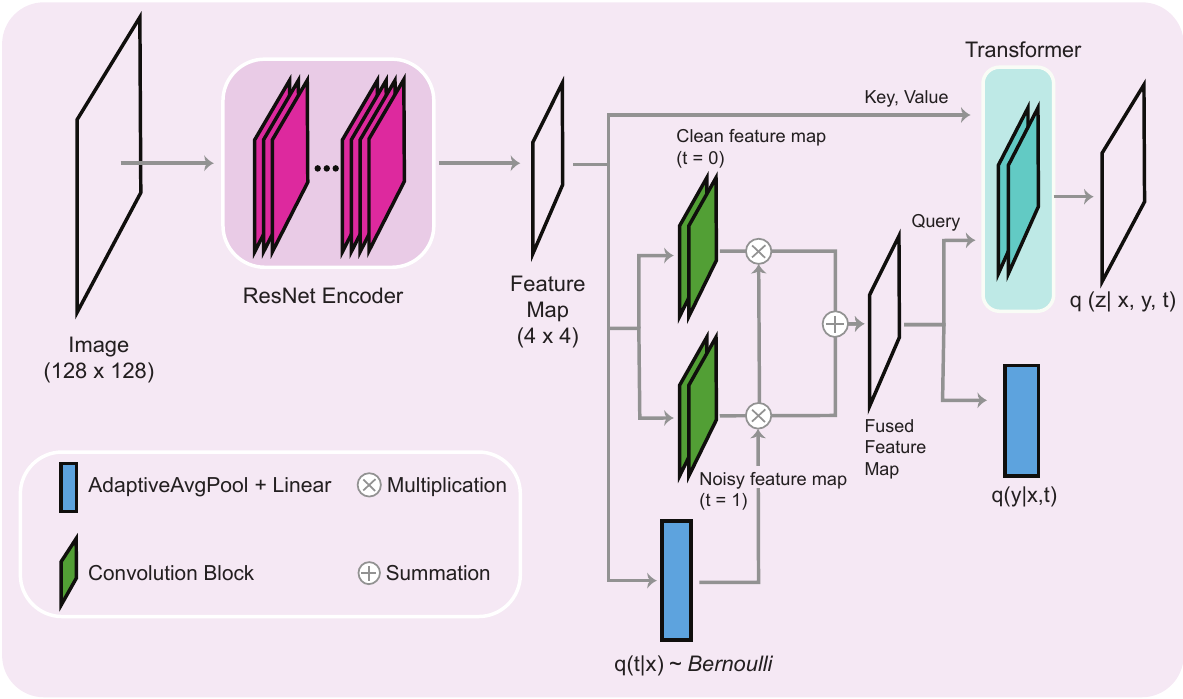}
\end{subfigure}
\quad
\begin{subfigure}{0.24\linewidth}
 \includegraphics[width=\linewidth]{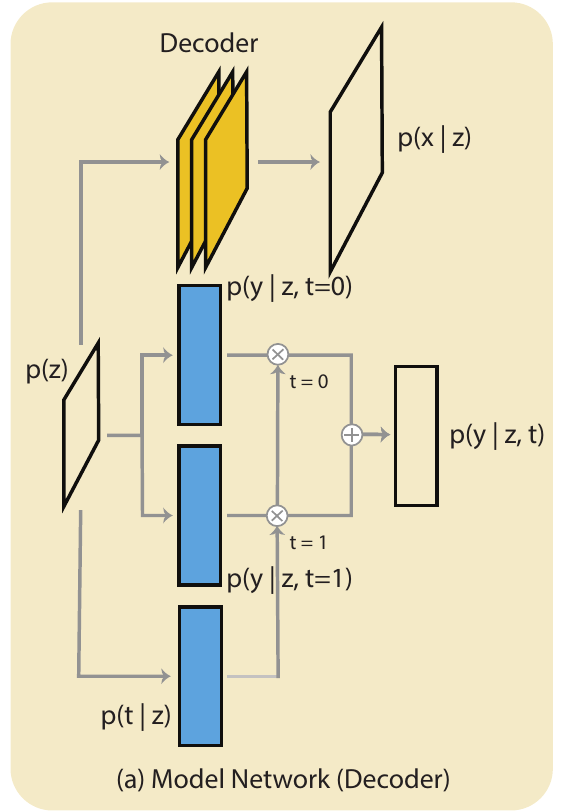}
\end{subfigure}
\end{center}
\vspace{-0.3cm}
\caption{The encoder (inference network) structure of our proposed causal transformer. We leverage bilinear fusion (BF) for $q(z|x,y,t)$ instead of concatenation~\citep{louizos2017causal}, and decoding conditional queries $H_z \sim q(y|x,t)$ and encoding features $H_x \sim p(x)$ as keys and values to conduct  attention. Decoder model is shown as Fig.~\ref{fig:network}~(a) with potential outcome modeling~\citep{imbens2010rubin, rubin1974estimating} from $p(z)$. } 
\label{fig:network}
\vspace{-0.4cm}
\end{figure*}

\subsection{Training Objective of TLT}

We build our TLT model based on the foundational framework of conditional variational encoder-decoder (CVED)~\citep{bahuleyan2018variational, kingma2013auto}, which learns a variational latent representation $z_i$ from data $x_i$ and conditional information (e.g., label $y_i$) for reconstruction or recognition.
To effectively learn visual causal pattern recognition, our TLT model uses variational inference to approximate the complex non-linear relationships involving: the pair probability ($p(x_i,z_i)$), the treatment likelihood $P(t_i)$, the model outcome $p(y_i)$, and the joint distribution $p(z_i, x_i, t_i, y_i)$. 
Specifically, we propose to characterize the causal graphical model in Fig. \ref{fig:figure1} \emph{(a)} as a latent variable model parameterized by a DNN encoder-decoder as shown in Fig. \ref{fig:figure2} (in Appendix A). Note that TLT uses an advanced decoding method $p(a_i)=F_{T}(H_x, H_z \sim P(x_i))$ for approximating $p(z_i)$ from $p(x_i)$ based on the  attention ($F_{T}$) from transformer~\citep{vaswani2017attention}, which will be detailed in Sec. \ref{section:att}. 

First, we assume the observations factorize conditioned on the latent variables and use an \emph{general} inference network (encoder) which follows a factorization of the true posterior. For the model network (decoder), instead of conditioning on observations, we approximate the latent variables $z$. 
For vision tasks, $x_i$ corresponds to a noisy input image indexed by $i$, $t_i \in \{0,1\}$
corresponds to the treatment assignment, $y_i$ corresponds
to the outcome and $z_i$
corresponds to the latent hidden confounder. Note that general formation of an approximation outcome ($\delta$) is modeling by $\delta_i=t_i\cdot y_i + (1-t_i)\cdot y_i$ as potential outcome model~\citep{imbens2010rubin, holland1986statistics} with its foundation over the causal inference. Next, each of the corresponding factors is described as:
    \begin{flalign*}
     & p(z_i) = \prod_{z\in z_i} \mathcal{N}(z | 0, 1) ;  ~~
     p(x_i | z_i) = \prod_{x \in x_i} p(x | z_i);  ~~
     p(t_i | z_i) = Bern(\sigma (f_1(z_i)));  \\
     & p(y_i | z_i, t_i) = \sigma(t_{i} f_{2}(z_i) + (1 - t_{i}) f_{3}(z_i) )
    \end{flalign*}    
with $\mathcal{N}(\mu, \sigma^2)$ denoting a Gaussian distribution with mean $\mu$ and variance $\sigma^2$, $p(x |z_i)$ being an appropriate probability distribution, $\sigma(.)$ being a logistic function, and $Bern(.)$ denotes the probability of success of a Bernoulli random variable. Each of the $f_k(.)$ function is an Adaptive Average Pooling plus Linear layer parameterized by its own parameters $\theta_k$ for $k=\{1, 2, 3\}$. Here $y_i$ is tailored for categorical classification problems, but our formulation can be naturally extended to different tasks. For example, one can simply remove the final $\sigma(.)$ layer of $p(y_i | z_i, t_i)$ for regression tasks.

Our TLT inference network (encoder), as illustrated in Fig. \ref{fig:network}, aims to learn meaningful causal representations in the latent space. As we can see from Fig. \ref{fig:figure1} \emph{(a)}, the true posterior over $z \in \mathbf{Z}$ depends on $x \in \mathbf{X}$, $t$, and $y$. We are required to know the treatment
assignment $t$ along with its outcome $y$ prior to inferring the distribution over $z$. Therefore, unlike variational encoders, which simply passes the feature map directly to latent space (the top path in our encoder), the feature map extracted from a residual block is provided to the other switching (the lower and middle paths in our encoder), which provides posterior estimates of treatment $t_i$ and outcome $y_i$. The switching mechanism (binary selection based on the treatment information of $t_i$ = 0 or 1) and its alternative loss training have been widely used in TARNet~\citep{shalit2017estimating} and CEVAE~\citep{louizos2017causal}
with theoretical and empirical justification. We employ the distribution by the switching mechanism:
    \begin{flalign}
    & q(t_i | x_i) = Bern(\sigma (g_1(x_i))); \\
    & q(y_i | x_i, t_i) = \sigma (t_{i}  g_{2}(x_i) + (1 - t_{i}) g_{3}(x_i) ),
    \label{eq:2}
    \end{flalign}
with each $g_k$ being a neural network approximating $q(t_i | x_i)$ or $q(y_i | x_i, t_i)$. They introduce auxiliary distributions that help us predict $t_i$ and $y_i$ for new samples. To optimize these two distributions, we  add an auxiliary objective to our overall model training objective over $N$ data samples:
{\small
    \begin{flalign}
    \mathcal{L}_{aux} = \sum^{N}_{i=1}( \log q(t_i=t_i^{*} | x_i^{*}) + \log q(y_i=y_i^{*} | x_i^{*}, t_i^{*})),
    \end{flalign} }%
 where $x_i^*$, $t_i^*$ and $y_i^*$ are the observed values in training set. Since the true posterior over $z$ depends on $x$, $t$ and $y$, finally we employ the posterior approximation below:
    \begin{flalign}
    q(z_i | x_i, y_i, t_i) = \prod_{z_i} \mathcal{N}(\bm{\mu}_i, \bm{\sigma^2}_i) 
    \label{eq:4}
    \end{flalign}
    \vspace{-0.6cm}
    \begin{flalign*}
    \bm{\mu}_i = t_i \bm{\mu}_{t=1, i} + (1 - t_i) \bm{\mu}_{t=0,i}, ~~
    \bm{\sigma^2}_i = t_i \bm{\sigma^2}_{t=1, i} + (1 - t_i) \bm{\sigma^2}_{t=0, i} 
    \end{flalign*}
    \vspace{-0.6cm}
    \begin{flalign*}
    \bm{\mu}_{t=0, i} = g_4 \circ g_0(x_i, y_i), \hspace{0.2cm} \bm{\sigma^2}_{t=0, i} = \sigma( g_5 \circ g_0(x_i, y_i)) \\
    \bm{\mu}_{t=1, i} = g_6 \circ g_0(x_i, y_i), \hspace{0.2cm} \bm{\sigma^2}_{t=1, i} = \sigma( g_7 \circ g_0(x_i, y_i)) 
    \end{flalign*}
where $g_k$ again denotes neural network approximation, and $g_0(x_i, y_i)$ is a shared, bilinear-fusioned representation of $x$, $t$ and $y$. More specifically, we multiply the feature map with approximated posterior $q(y_i|x_i, t_i)$ without logistic function $\sigma$ to get $g_0(x_i, y_i)$. Finally, we can have the overall training objective for the inference and model networks.
The variational lower bound of TLT to be optimized is given by:
    {\small
    \begin{flalign}
    &\mathcal{L}_{TLT} = \mathcal{L}_{aux} + \sum^{N}_{i=1}  \mathbb{E}_{q(z_{i}|x_{i}, t_i, y_i)}[  \log p(x_i, t_i | z_i) + \log p(y_i|t_i, z_i) 
     + \log p(z_i) - \log q(z_i|x_i, t_i, y_i)]. 
    \end{flalign}
       }%
As shown in Fig.~\ref{fig:figure2} (in Appendix A), we could model $q(t|x)\doteq p(t)$ to access the treatment information directly for training to guide one corresponding sub-network in Fig.~\ref{fig:network}; for testing, $q(t|x)$ could be inferred by a given input $x$ without knowing treatment information from an unsupervised perspective. 

\subsection{Attention mechanism of TLT}
\label{section:att}
Attention mechanism is one of the human learning components to capture global dependencies for discovering logical and causal relationships~\citep{nauta2019causal} from visual patterns in the cognitive psychology community~\citep{chen2015role}. Transformer~\citep{vaswani2017attention} based attention mechanism has, recently, shown its connection from the \textbf{sequential} energy update rule to Hopfield networks~\citep{ramsauer2020hopfield}, which stands for a major framework to model human memory. With the intuition on leveraging human-inspired attention upon inference from noisy images, we incorporate a new type of Transformer module for the proposed causal modeling, 
which explicitly model all pairwise interactions between elements in a sequence. The idea is to learn the causal signal~\citep{lopez2017discovering} via self-attention setup, where we set the interference signal ($H_z$) for learning query and image features ($H_x$) for learning key and value. As shown in Fig~\ref{fig:network}, we use a feature map with a ResNet$_{34}$~\citep{he2016deep} encoder extracting from input image $p(x_i)$ feeding into keys ($K$) and value ($V$)  with queries $q(y_i)$ from Eq. (\ref{eq:2}):
\begin{align}
    Q =&~\operatorname{unroll}\left(F_Q( H_z \sim q(y_i | x_i, t_i))\right) ;~~
    K =\operatorname{unroll}\left(F_K( H_x \sim p(x_i))\right) \\
     V =&~\operatorname{unroll}\left(F_V( H_x \sim p(x_i))\right); ~~
    a_i =\operatorname{softmax}\left(\frac{Q K^{T}}{\sqrt{d_{k}}}\right) V
\end{align}
where $F_Q$, $F_K$, $F_V$ are convolutional neural networks and $d_{k}$ is dimension of keys. Finally, we model $q(z_i)$ by using $q(t_i|x_i)$ and $p(a_i|x_i)$ with the causal two model extended from Eq. (\ref{eq:4}) for approximating posterior distribution $p(z_i)$:
\begin{flalign}
    p(z_i)\leftarrow q(z_i | x_i, a_i, y_i, t_i) = \prod_{z_i} \mathcal{N}(\bm{\mu}_i, \bm{\sigma^2}_i).
    \end{flalign}
We also have conducted ablation studies on architecture selection and required parameters with respect to supervised learning~\citep{he2016deep}, attention networks~\citep{vaswani2017attention}, and causal model~\citep{shalit2017estimating} in \textbf{supplement}~\ref{sup:sect:aba} to validate our model design of TLT. To sum up, the proposed causal architecture attains the best performance with the same amount of parameters.

\section{Evaluating Causal Effects on Noisy Images}

In this section, we introduce noisy image datasets and conduct statistical refutation tests on TLT to evaluate its causal effect based on the CGM in Fig. \ref{fig:figure1} \emph{(a)}. That is, we provide an affirmative answer to whether there \textbf{exist causal effects} in the studied noisy image classification tasks.

\subsection{Estimate Causal Effects}

Estimation of \textbf{expected causal effects} is one general approach~\citep{pearl2019seven, pearl2009causality, louizos2017causal} to evaluate whether a CGM (from a logic hypothesis) is valid on the selected test dataset. The underlying graphical model will undergo a series of randomization tests of graphical connection and sub-set sampling to measure its estimation errors on estimating causal effects. In general, a causal model is reliable with the CGM when exhibiting a lower absolute error on the causal effects. In this work, we use average treatment effects (ATE), as used in prior arts~\citep{louizos2017causal}, for comprehensive analysis.\\
\textbf{Average Treatment Effects (ATEs).}
In the binary treatment setting~\citep{pearl1995causal}, for the $i$-th individual and its associated model outcome $y_i$ considering the treatment effect, the ATE is calculated by:
\begin{align}
 y_{i}&=y_{t_{i}=0,i}\left(1-t_{i}\right)+y_{t_{i}=1,i} (t_{i}), \\
 ATE&=|\mathbb{E}\left[y_{i}=y^{*}_{i}|t^{*}_{i}=1\right]-\mathbb{E}\left[y_{i}=y^{*}_{i}|t^{*}_{i}=0\right]|,
\end{align}
where $y_{t_{i},i}$ denotes the prediction with estimated treatment $t_i \in \{0,1\}$. $y^{*}_{i}$ and $t^{*}_{i}$ are the observations. The ATE is taken over all subjects.
From \citep{greenland1999confounding}, these metrics cannot be properly estimated if there are confounding variables in the system. On the other hand, Pearl \citep{pearl1995causal} introduces the ``do-operator''~\citep{pearl2019seven} on treatment to study this problem under intervention. The $do$ symbol removes the treatment $t$ from the given mechanism and sets it to a specific value by some external intervention. The notation $P(y|do(t))$ denotes the probability of $y$ with possible interventions on treatment. Following Pearl’s back-door adjustment formula \citep{pearl2009causality} and the CGM in Fig. \ref{fig:figure1}, it is proved in \citep{louizos2017causal} that the causal effect for a given binary treatment $t$, a proxy variable $x$, an outcome $y$ and a confounding variable $z$ can be evaluated by (similarly for $t=0$):
\begin{equation}
\begin{array}{l}{p(y | x, d o(t=1))}= {\int_{z} p(y|x, t=1, z) p(z | x) d z}\end{array}
\label{eq:eq:p:do}
\end{equation}

To intervene the information of $t$ ($do(t)$), flipping errors~\citep{louizos2017causal} with different rates 
(see \textbf{supplement} \ref{sup:arch:sec}) are applied to change the $t_i$ label(s) \citep{paszke2017automatic} in our experiments in Section \ref{subsec_test_causal}. The proposed CGM and its associated TLT show resilient ATE estimation under statistical refutations.\\

\textbf{Visual Patterns in the Intervention Level ($\mathbb{II}$).}
We clarify two common scenarios, noisy context and under perturbation, in the intervention level ($\mathbb{II}$) for noisy image classification. As shown in Tab.~\ref{tab:2:do:x}, the treatment information ($t$) is binary with an accessible noisy input $x$ and \textbf{inaccessible} ontological (clean) representation $\tilde x$ from Eq.~(\ref{eq:c:h:do}) for visual pattern modeling. Next, we introduce datasets in the regime of the case 1 and 2 for our experiments in this work.

\begin{table}
\caption{Applying causal modeling on noisy images classification. \textbf{Case 1}: perfect labeled visual pattern (H$_{lab}$) with secondary patterns (H$_{iid}$); \textbf{Case 2}: original labeled images (H$_{ori}$) under an additive perturbation (F$_{per}$). }
\label{tab:2:do:x}
\centering
\begin{tabular}{|l|c|c|c|}
\hline
 \textbf{Treatment} & \textbf{$\tilde x$} & \textbf{$do(\tilde x)$} & \textbf{$t$}=1 or 0  \\ \hline
\textbf{1.} Context & H$_{lab}$ & H$_{lab}$+ H$_{iid}$ & Additional patterns (e.g., ``waterside'') (1) or not (0) \\ \hline
\textbf{2.} Perturbation & H$_{ori}$ & F$_{per}$(H$_{ori}$) & Artificial noise (e.g., Gaussian) (1) or not (0) \\ \hline
\end{tabular}
\vspace{-2mm}
\end{table}

\subsection{Case 1: NICO Dataset with Noisy Extra Visual Patterns}
NICO~\citep{he2020towards} is a large-scale and open-access benchmark dataset for noisy image classification, which is motivated by studying non-independent image classification with causal modeling. The NICO dataset labels images with both main concepts (e.g., ``cat'') and contexts as sub-labels (e.g., ``water''). NICO is constructed by two super-classes: ``animal'' and ``vehicle'', with 10 classes for ``animal'' and 9 classes for ``vehicle''. In total, NICO contains 19 classes, 188 contexts, and 25,000 images. The design intuition of NICO is to provide a causal modeling benchmark for large-scale image classification. The authors evaluate several major image classification dataset (e.g., ImageNet, Pascal, and MS-COCO) and found out the auxiliary context information (treatment) is much random and inaccurate from statistical measurement for structuring validated causal inference. By selecting different contexts of the concept, testing data distribution can be unknown and different from training data distribution, which can be used to evaluate a causal inference model.

In our experiments, we follow the standard NICO evaluation process~\citep{he2020towards}, where a concept is incorporated with two contexts. We further use \textbf{context} as treatment in the intervention level as in \textbf{Case 1} of~Tab.~\ref{tab:2:do:x}. One context is the attribute of concept ($t=1$) while another context is the background or scene of a concept ($t=0$). 

\subsection{Case 2: Curated Causal Pairs (CPS) Dataset with Additive Artificial Noises}
Despite many efforts in providing benchmark datasets for causal inference on non-vision tasks \citep{hill2011bayesian, lalonde1986evaluating, hoyer2009nonlinear, peters2014causal}, visual causal data collection is relatively limited to bare causal effect evaluation with conditional visual treatments \citep{lopez2017discovering}. Motivated by the perturbation-based causation studies testing biological network and the efforts from NICO, we further curate two datasets from public sources, named causal pairs \textbf{(CPS)}, by using a diverse set of image perturbation types as treatment (i.e., \textbf{Case 2} in Tab.~\ref{tab:2:do:x}). 

We select two representative datasets, Microsoft COCO \citep{lin2014microsoft}, and a medicine dataset, Decathlon \citep{simpson2019large}, to create our CPS datasets. Each CPS contains pairs of original and perturbed images, as well as \textbf{five} different perturbation types described in Sec. \ref{subsec_type}.
Table.~\ref{tab:3:dataset} summarizes the NICO and our CPS datasets. Next, we introduce how to generate noisy images in CPS. 

\begin{figure}[ht!] %
\centering
\begin{subfigure}[b]{0.321\textwidth}
  \includegraphics[width=\textwidth]{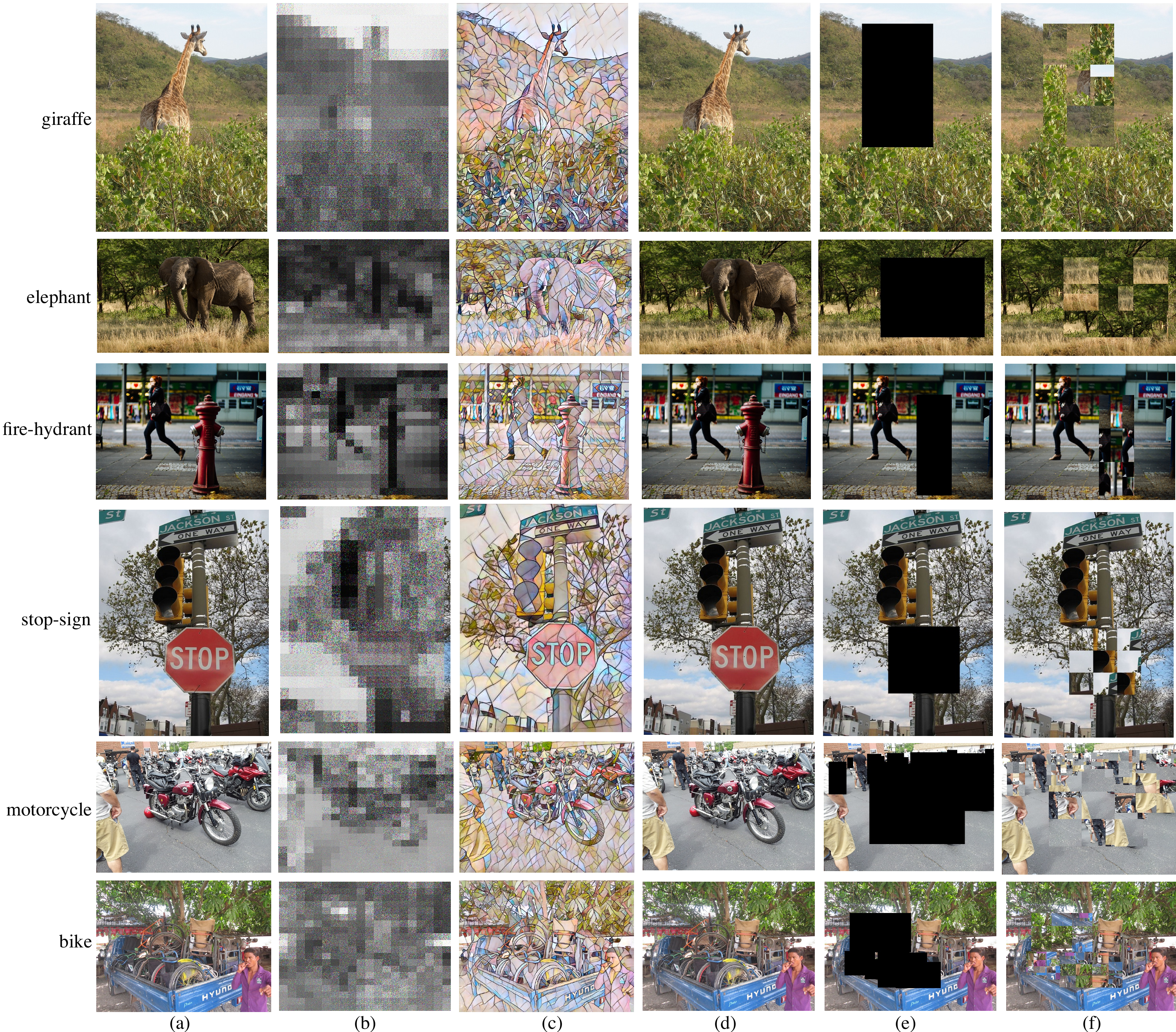}
\vspace{-0.1cm}
\label{fig:figure3}
\end{subfigure}
\begin{subfigure}[b]{0.251\textwidth}
        \includegraphics[width=\textwidth]{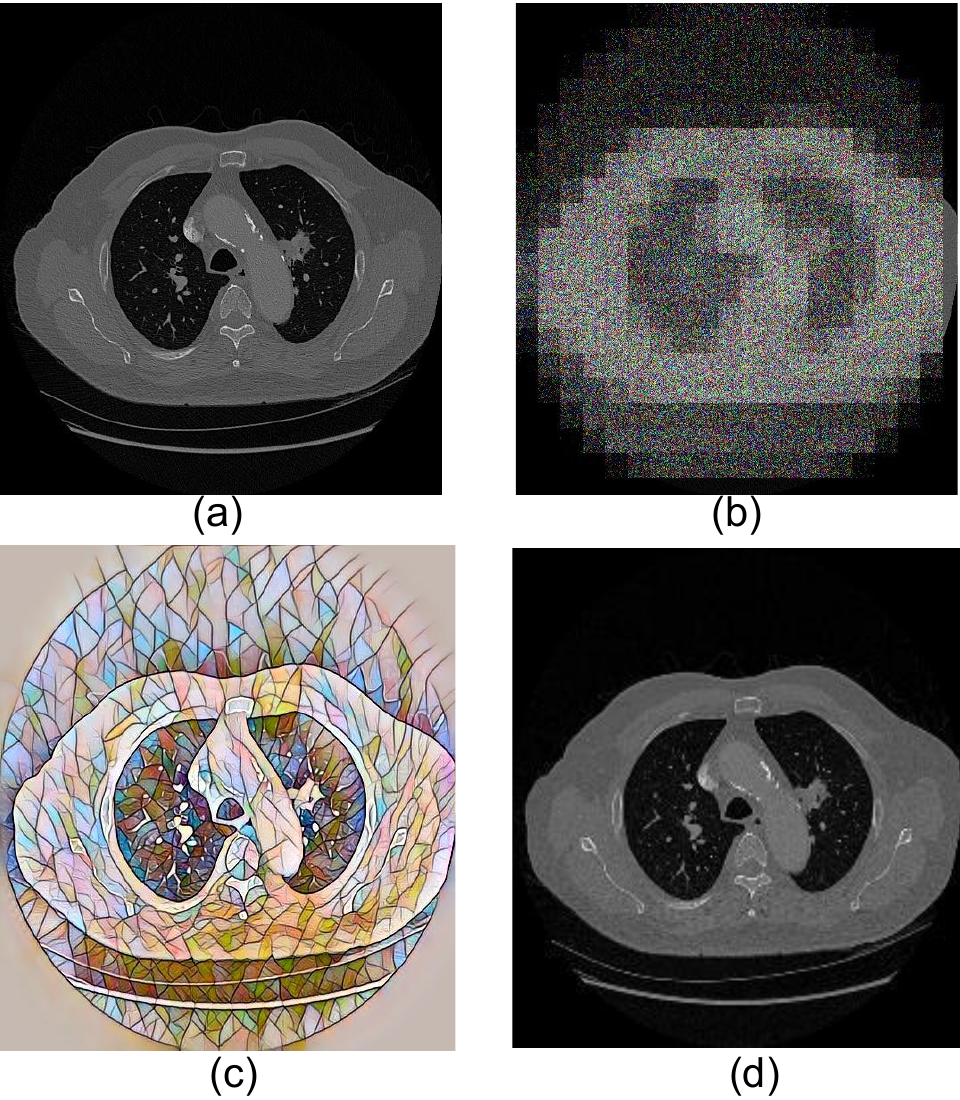}
        \vspace{-0.2cm}
        \label{fig:tiger}
    \end{subfigure}
          \caption{Illustration of our generated \textbf{CPS} dataset for noisy image classification. We randomly selected $\mathbf{50\%}$ of labeled images from both datasets and applied visual modifications to generate interventional observations. We selected similar object classes by 1,000 human surveys. \textit{Left}: three causal pairs -- giraffe/elephant, fire-hydrant/stop-sign, and motorcycle/bike. From left to right in \textbf{CPS}, the visual treatments are: (a) original input image, (b) image scrambling, (c) neural style transfer; (d) adversarial example.  We further discuss masking intervention effects used in~\citep{lopez2017discovering, yang2019causal} on general subjects by (e) object masking; and (f) background refilling. \textit{Right}: a demonstration of Lung Tumours in Decathlon of the same format.} 
\vspace{-0.2cm}
\label{fig:figure:7:ok}
\end{figure}

\begin{table}[ht!]
\centering
\caption{Comparison of noisy image classification datasets: CPS with perturbation as a treatment (see Sec. \ref{subsec_type}) and NICO with noisy context (e.g., ``indoor'').}
\label{tab:3:dataset}
\begin{adjustbox}{width=0.88\textwidth}
\begin{tabular}{|l|l|c|c|c|}
\hline
\textbf{Dataset} & \textbf{Treatment (Binary Information)} & \textbf{Numbers} & \textbf{Super-classes} & \textbf{Total classes} \\ \hline
CPS (ours) & Receiving artificial noise (or not) & 13,752 & General / Medical & 16 \\ \hline
NICO & Existing context-wise pattern (or not) & 25,000 & Animal / Vehicle & 19 \\ \hline
\end{tabular}
\end{adjustbox}
\vspace{-2mm}
\end{table}

\textbf{Super-class 1: Generating Noisy General Objects.}
To generate CPS dataset from MS-COCO~\citep{lin2014microsoft} for general super-class, we selected \textbf{six }similar object classes that could possibly result in confusing interpretation and recognition by human psychology studies~\citep{reed2012cognition, musen1990implicit} (e.g., giraffe and elephant, etc.). We conduct a survey with 1,000 volunteers from Amazon mechanical turk~\citep{turk2012amazon} and pick the top-3 similarity label pairs. Specifically, we format three different common causal pairs, namely giraffe-elephant (CPS$_1$)  with 3316 images, stop sign-fire hydrant (CPS$_2$) with 2419 images, and bike-motorcycle (CPS$_3$) with 4729 images, where the dataset is visualized in Fig.~\ref{fig:figure:7:ok} (a). \\
\textbf{Super-class 2: Generating Noisy Medical Images.}
For the medical super-class, we use an identical setting with 2630 training and 658 test CT images for \textbf{ten} different types (total classes) of human disease from Decathlon \citep{simpson2019large}, which includes: (1) Liver Tumours; (2) Brain Tumours; (3) Hippocampus; (4) Lung Tumours; (4) Prostate; (5) Cardiac; (6) Pancreas Tumour; (7) Colon Cancer; (8) Hepatic Vessels, and (10) Spleen. More details and visualization (Fig.~\ref{fig:figure:7:ok} (b)) about this dataset are given in \textbf{supplement} \ref{sup:sect:aba}. From these two super-classes, we randomly selected $\mathbf{50\%}$ of these labeled images and applied visual modifications to generate interventional observations. Each generated image is assigned with a  binary treatment indicator vector $t_i$, where its $i$-th element denotes the binary treatment label according to the $i$-th visual modification. 

\subsection{Visual Perturbation (Treatment) in CPS}
\label{subsec_type}
We employ five distinct types of image modification methods as independent intervention variables: (i) image scrambling; (ii) neural style transfer; (iii) adversarial example; (iv) object masking, and (v) object-segment background shifting. Below we provide brief descriptions for these visual treatments as illustration in Fig.~\ref{fig:figure:7:ok}.\textbf{Image Scrambling (IS)} \citep{ye2010image} algorithms re-align all pixels in an image to different positions to permute an original image into a new image, which is used in privacy-preserved classification~\citep{tarr1998image}. 

\textbf{Neural Style Transfer (ST)} ~\citep{gatys2015neural} creates texture effect with perceptual loss~\citep{johnson2016perceptual} and super-resolution along with instance normalization~\citep{ulyanov2016instance}. 

\textbf{Adversarial Example (AE)}
adds input perturbation for prediction evasion. We employ the Fast Gradient Sign Method (FGSM) \citep{goodfellow2015laceyella} with a scaled $\ell_{\infty}$ perturbation bound of $\epsilon = 0.3$. We also evaluated other attacks including C\&W~\citep{carlini2017adversarial} and PGD~\citep{madry2017towards} in \textbf{supplement} ~\ref{sup:sect:aba}. 

\textbf{Object Masking (OM) \& Background Refilling (BR)}: 
Object masking (OM) was proposed in previous studies \citep{lopez2017discovering, yang2019causal} for causal learning. We applied OM and another masking methods, background refilling (BR), that duplicates non-object background into the mask segment as treatments. 

\section{Experiments}
\subsection{Noisy Image Classification on NICO and CPS}
\begin{table*}[t]
\centering
\caption{Perturbation (e.g., texture) effects with classification accuracy (\%) on the average of \textbf{CPS} images ($\sim$13.7k) for different treatments and their causal effect estimates. 
Note that TLT (7.39M) has similar parameters compared with CVAE' and CEVAE', which are enhanced by ResNet as discussion in the ablation studies. \textbf{n} is for treatment noise level. }
\label{tab:texture}
\vspace{-2mm}
\adjustbox{max width=0.85\textwidth}{
\begin{tabular}{l|lll|lll}
\cline{2-7}
 & \multicolumn{3}{l|}{\textbf{Classification Accuracy} ($\uparrow$)} & \multicolumn{3}{l|}{\textbf{Average Treatment Effect} ($\uparrow$)} \\ \hline
\multicolumn{1}{|l|}{\textbf{Type of t (with n = 0.05)}} & \multicolumn{1}{l|}{CVAE'} & \multicolumn{1}{l|}{CEVAE'} & TLT (ours) & \multicolumn{1}{l|}{CVAE'} & \multicolumn{1}{l|}{CEVAE'} & \multicolumn{1}{l|}{TLT (ours)} \\ \hline
\multicolumn{1}{|l|}{Original (without $\mathbf{t}$)} & 83.31 $\pm0.12$ & 83.31 $\pm0.23$ & 83.31 $\pm0.13$ & 0.012 & 0.018 & \multicolumn{1}{l|}{\textbf{0.032}} \\ \hline
\multicolumn{1}{|l|}{Style Transfer (ST)} & 73.67 $\pm0.31$ & 74.34 $\pm0.26$ & \textbf{76.12} $\pm0.27$ & \textbf{0.354} & 0.343 & \multicolumn{1}{l|}{0.318} \\ \hline
\multicolumn{1}{|l|}{Image Scrambling (IS)} & 72.31 $\pm1.27$& 76.21 $\pm0.81$ & \textbf{80.12} $\pm0.54$ & 0.057 & \textbf{0.295} & \multicolumn{1}{l|}{0.288} \\ \hline
\multicolumn{1}{|l|}{Adversarial Example (AE)} & 79.12 $\pm0.25$& 81.12 $\pm0.17$& \textbf{83.12} $\pm0.12$ & 0.025 & 0.027 & \multicolumn{1}{l|}{\textbf{0.036}} \\ \hline
\multicolumn{1}{|l|}{Object Masking (OM)} & 70.12 $\pm0.19$& 72.73 $\pm0.21$& \textbf{73.06} $\pm0.11$& 0.179 & 0.241 & \multicolumn{1}{l|}{\textbf{0.243}} \\ \hline
\multicolumn{1}{|l|}{Background Refilling (BR)} & 71.32 $\pm0.28$& 72.59 $\pm0.29$& \textbf{73.91}$\pm0.17$ & 0.213 & 0.221 & \multicolumn{1}{l|}{\textbf{0.238}} \\ \hline
\end{tabular}
}
\vspace{-2mm}
\end{table*}

\textbf{Generative Model Baselines}

For a fair comparison, we select two benchmark conditional generative model incorporating both information of label ($y$) and binary treatment ($t$): modified conditional VAE~\citep{sohn2015learning, kingma2013auto} (CVAE') and modified CEVAE~\citep{louizos2017causal} (CEVAE'), where CVAE' use $p(t,y)$ for concatenation as a conditional inference and CEVAE' follows a similar causal variational inference process~\citep{louizos2017causal} without features fusion and conditional queries. Both model are enhanced by ResNet~\citep{he2016deep} and attention layers with similar parameters (7.1M) with TLT. 
Noted CEVAE~\citep{louizos2017causal} is originally designed and applied only on linear regression tasks but benefited from our causal modeling for noisy image classification.

\begin{wraptable}{r}{0.5\textwidth}
\caption{Classification accuracy (\%) on NICO.}
\label{tab:nico:acc}
\centering
\begin{adjustbox}{width=0.48\textwidth}
\begin{tabular}{|l|c|c|c|c|}
\hline
\textbf{Model} & \textbf{StableNet}~\citep{he2020towards} & \textbf{CVAE'} & \textbf{CEVAE'} & \textbf{TLT} \\ \hline
Acc. & 59.76 $\pm1.52$ & 57.23 $\pm2.12$ & 62.17 $\pm1.82$ & \textbf{65.98} $\pm1.74$ \\ \hline
\end{tabular}
\end{adjustbox}
\vspace{-2mm}
\end{wraptable}

\textbf{Performance on NICO Dataset.} 
We first evaluate models performance trained on NICO dataset. From the reported results in the paper~\citep{he2020towards, zhang2021deep}, we select the best reported model, StableNet from~\citep{zhang2021deep} with sample weighting. As shown in Table.~\ref{tab:nico:acc}, generative models with proposed causal modeling attain competitive results on NICO with compositional bias setup, where TLT attains a best performance of \textbf{65.98}\%. We provide more analysis under different setup of NICO, where TLT remains as the best model in \textbf{supplement}~\ref{sup:arch:sec}. 

\begin{figure}
\begin{center}
\begin{subfigure}{0.35\textwidth}
    \includegraphics[width=\textwidth]{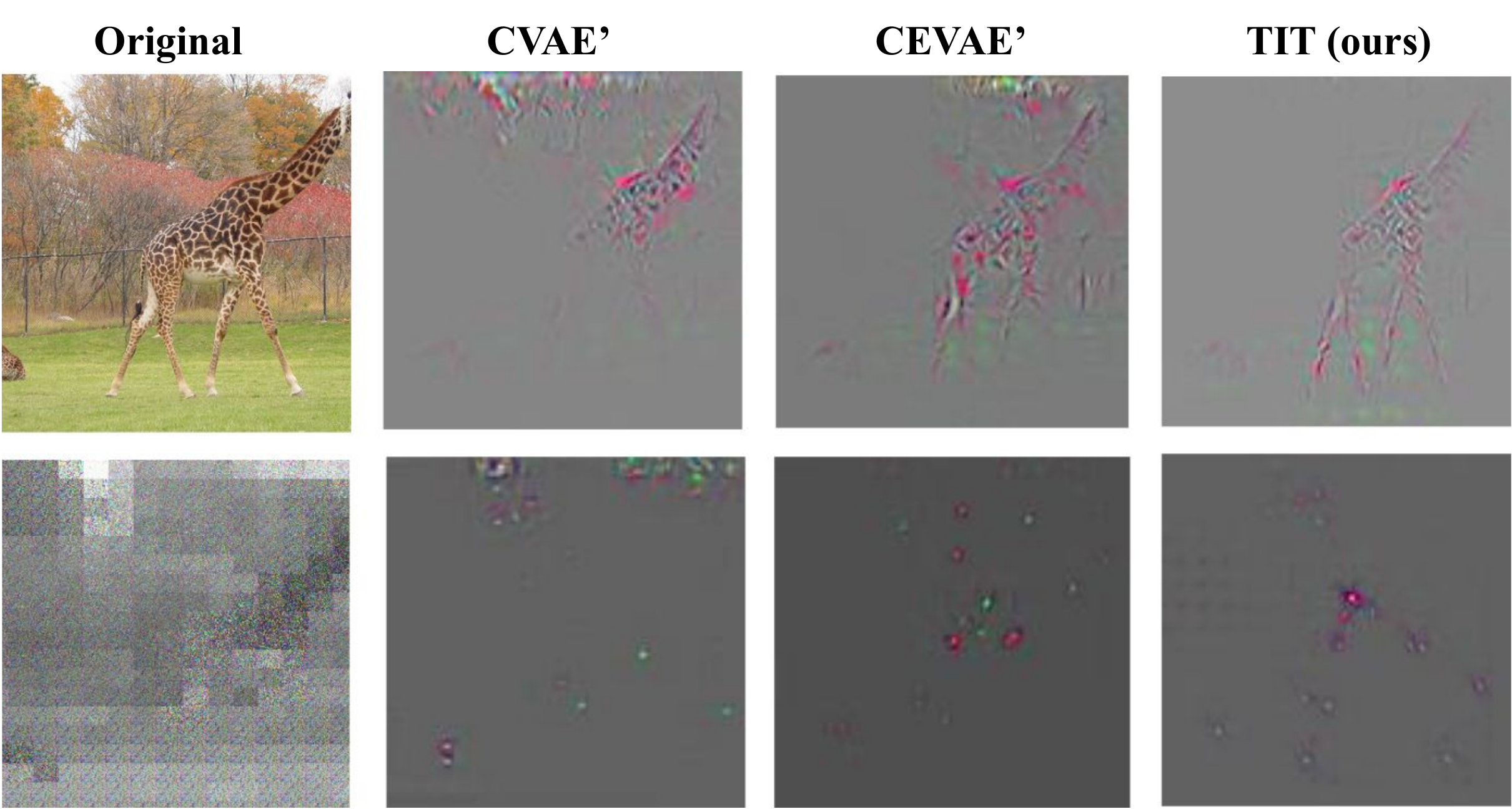}
  \end{subfigure}
  \quad
  \begin{subfigure}{0.40\textwidth}
    \includegraphics[width=\textwidth]{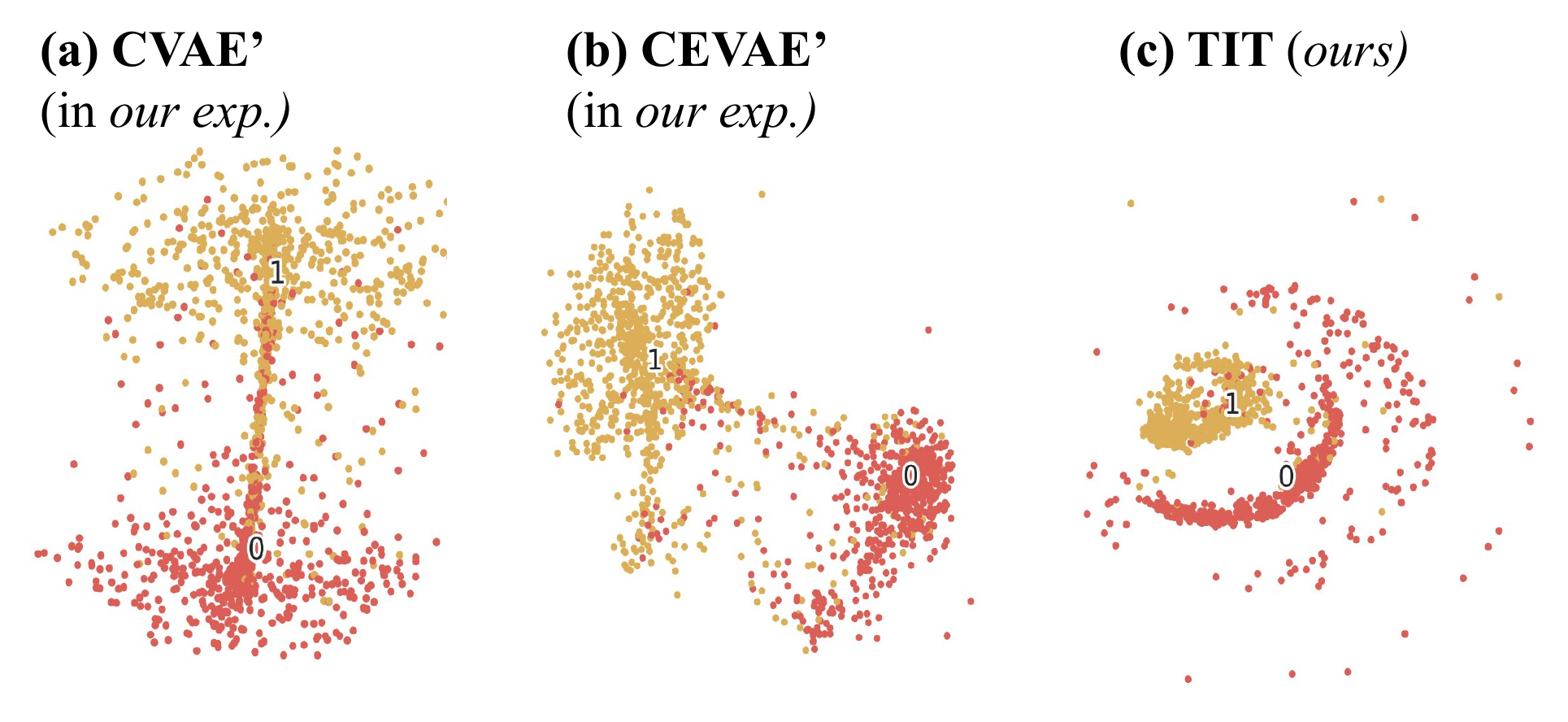}
  \end{subfigure}
  \end{center}
  \vspace{-4mm}
\caption{\emph{(a)} With proposed TLT and CPS dataset, neural saliency methods can be extended to visual pattern from inference. Take the top row as an example, using TLT, guided grad-CAM \citep{selvaraju2017grad} can be more aligned with the concise human-interpretable giraffe patterns instead of forest texture and edges. More correlation analyses between saliency and labels in NICO and CPS are given in supplement~\ref{sup:arch:sec}. \emph{(b)} Visualization of learned manifolds of $q(z)$ by tSNE~\citep{maaten2008visualizing}, proposed CTR’s results largest intra-cluster pair-wise sample distances between 
additive noise (adversarial) ($t$ = 1) and vanilla ($t$ = 0) image samples from CPS$_1$.} 
\label{fig:figure:6}
\vspace{-2mm}
\end{figure}

\textbf{Performance on CPS Dataset.}
In Table \ref{tab:texture}, we compare TLT with modified CVAE' and modified CEVAE' as baselines  trained on \textbf{CPS} dataset. The accuracy of TLT in the original image, IS, ST, AE, OM and BR settings are consistently better than CVAE' and CEVAE', with substantially large margins ranging from $\mathbf{1.60\%}$ to $\mathbf{7.81\%}$. CEVAE' and TLT are also shown to have higher causal estimate (CE) than CVAE' in all settings except for ST. Interestingly, ST leads to a higher causal value (from $\mathbf{0.318}$ to $\mathbf{0.354}$) when compared to the other modifications such as IS and AT. This finding accords to the recent studies on DNN's innate bias of using edges and textures for vision task \citep{geirhos2018imagenet}. CEVAE' and TLT having lower value in ST setting could be explained by a more unbiased representation learned by inference network with lower dependency on edges and textures. A benchmark visualization of Guided Grad-CAM \citep{selvaraju2017grad} in Fig. \ref{fig:figure:6} \emph{(a)} validates this hypothesis and highlights the importance of our inference network in gaining robust visual understanding from latent space $z$ as tSNE~\citep{maaten2008visualizing} results  Fig.~\ref{fig:figure:6} \emph{(6)}. One critical issue for visual intervention is its difficulty in investigating the effect on object mask size \citep{lopez2017discovering, pickup2014seeing}. \textbf{supplement~\ref{sup:arch:sec}} shows a consistent and stable performance of TLT against varying mask sizes.

\textbf{Case Study on the medical super-class:} We conduct the same experiments with medical super-class to identify visual clinical features. Both the classification and estimation performance are consistent with general CPS objects, where TLT attains the highest accuracy $88.74$\% in the original setting and $82.57$\%  in the scrambling setting (e.g., data encryption operation) settings. TLT is most effective in classifying noisy image and more sensible in measuring ATE on adversarial example. We also conduct expert evaluation on the activation saliency of clinical patterns (Fig. \ref{fig:figure:6}). Based on their domain knowledge \citep{wang2017chestxray,rajpurkar2017chexnet,rajpurkar2017mura}, three physicians independently and unanimously give the highest confidence scores on saliency attributes to our method. 

\textbf{Statistical Refutation of Causal Models:}
\label{subsec_test_causal}
To rigorously validate our ATE estimation result, we follow a standard refuting setting \citep{rothman2005causation, pearl2016causal,pearl1995testability} with the causal model in Fig. \ref{fig:figure1} to run three major tests, as reported in \textbf{supplement}~\ref{sup:sec:refu} and Table S\ref{tab:noise}, which validate our method is robust.

\vspace{-2mm}
\section{Conclusion}
Motivated by human-inspired attention mechanism and causal hierarchy theorem, in this paper
we proposed a novel framework named treatment learning transformer (TLT) for tackling noisy image classification with treatment estimation. In addition to showing significantly improved accuracy of TLT on the NICO dataset with noisy contexts, we also curated a new causal-pair dataset (CPS) based on five different visual image perturbation types for performance benchmarking on general and medical images. We validated the causal effect of TLT through statistical refutation testing on average treatment effects. We also show derived advantages of TLT in terms of improved visual saliency maps and representation learning. Our results suggest promising means and a new neural network architecture toward the advancement of research in the intersection of deep learning and visual causal inference.

\bibliography{ref}
\bibliographystyle{unsrtnat}

\clearpage
\appendix
\section{The ``What If'' Challenges for Deep Neural Networks (DNNs)}
\label{sup:causal}
A causal inferable DNNs for reasoning chaotic real-world patterns~\citep{pearl2009causality, yang2019causal, pearl1995causal} would be necessary for many practice scenarios. 
Recent regulatory concerns (e.g., GDPR \citep{albrecht2016gdpr}) on Artificial Intelligence (AI) safety and self-driving automobile accidents also highlight the importance and the emergence of understanding: (1) "What" does a DNNs model "learn for accurate label-prediction and (2) utilizing the "Why" relationship between labels generated by human knowledge and their conceptional pattern representations in the real world. 

We propose COCO$_{CP}$ dataset based on MS-COCO~\citep{lin2014microsoft}. COCO$_{CP}$ dataset includes classes having similar object(concept) but are  different in context that could possibly result in confusing interpretation and recognition by human psychology studies (e.g., giraffe and elephant, etc.). In total, three different common causal pairs are formatted, namely giraffe-elephant (g-e)  with 3316 images, stop sign-fire hydrant (s-f) with 2419 images, and bike-motorcycle (b-m) with 4729 images.   A pair of stop sign and fire hydrant has been selected to study public awareness of on-road visual detection. 

\subsection{Neural Causation Coefficient (NCC)}
Neural Causation Coefficient (NCC)~\citep{lopez2017discovering} is a novel observational causal discovery technique for the joint distribution of a pair of related proxy variables that are computed by applying CNNs to the image pixels. NCC leverages such embedded joint distribution as a regularization term to encourage the learning of causal or anticausal patterns in neural networks. Lopez et al.~\citep{lopez2017discovering} used an augmented NCC network to prove the existence of causal relations in ResNet~\citep{he2016deep} between object and context in an image, and showed that in object-feature ratio anticausal signal consistently has stronger relation than causal signal. However, they mainly focus on discussing the object-context causal hypothesis in the image setting, but barely covers the measurement of perturbation effect.

\begin{figure} [ht!]
\centering
\begin{subfigure}{\linewidth}
   \centering
   \includegraphics[width=\linewidth]{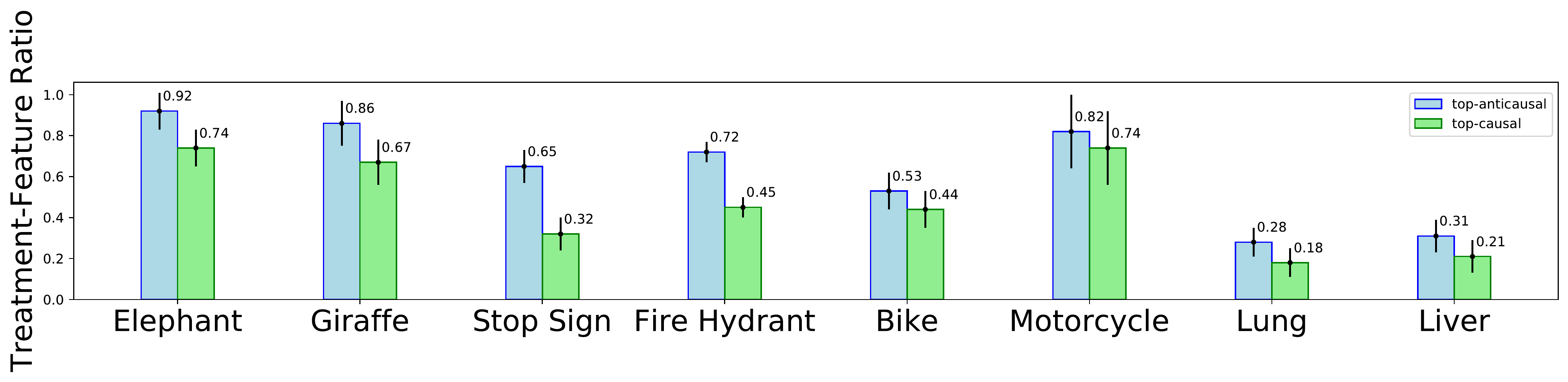}
   \caption{TFR calculated by feature f$_{R}$ from ResNet34~\citep{he2016deep} as \citep{lopez2017discovering}.}
   \label{fig:ncc:1} 
\end{subfigure}
\begin{subfigure}{\linewidth}
   \centering
   \includegraphics[width=\linewidth]{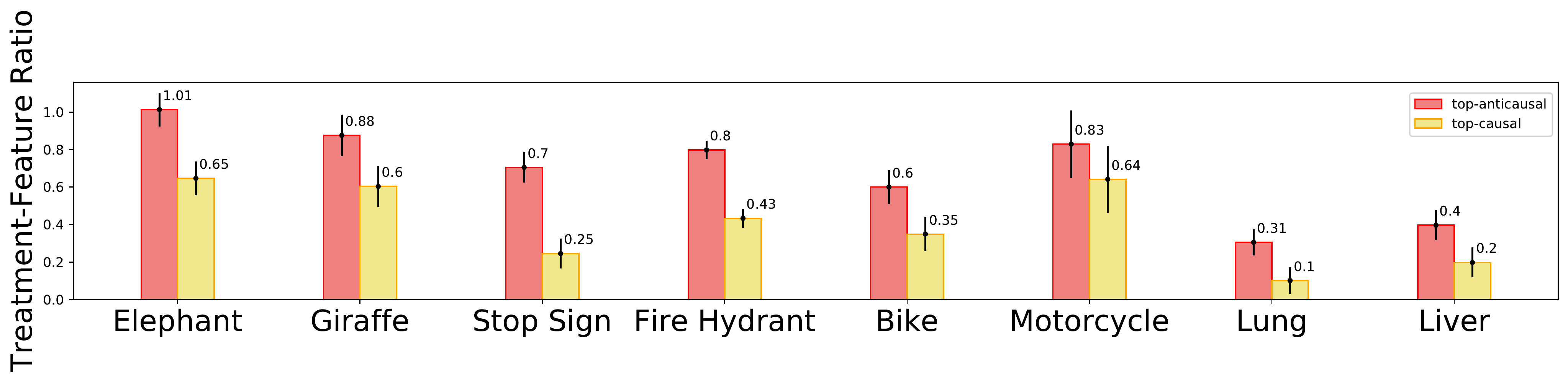}
   \caption{TFR calculated by feature f$_{C}$ from our proposed CAN. }
   \label{fig:ncc:2}
\end{subfigure}
\vspace{-0.3cm}
\caption{Evaluation of causal pairs by treatment feature ratio (TFR) score \citep{lopez2017discovering}. The average and standard deviation of TFR associated to the top-1\% causal/anticausal feature scores are displayed.
The results show the visual perturbation measurement is coherent with the previous study~\citep{lopez2017discovering}.} 
\label{fig:figure:5:ncc}
\end{figure} %

From Eq. \ref{eq:eq:p:do}, \textbf{object-feature ratio (OFR)}~\citep{lopez2017discovering} could be extended to a \textbf{treatment-feature ratio (TFR)} score s$^{t}_{l}$ by do-operator~\citep{ pearl2019seven} as:

\begin{equation}
s_{l}^{t}=\frac{\sum_{j=1}^{m}\left|(f_{j l}^{c}|do(t))-(f_{j l}|do(t))\right|}{\sum_{j=1}^{m}\left|(f_{j l)}|do(t)\right|},
\end{equation}
where ($l$=512) is the input feature length, and f$^{t}_{j}$ corresponding vectors of feature scores is equivalent to F$_{\text{DNN}}$(x$^{t}_{j}$). 

We reproduce the NCC architecture from ~\citep{lopez2017discovering} and find all the anti-causal scores of COCO is larger than causal score as shown as~\citep{lopez2017discovering}. 
\subsection{Correlation and Causation in Vision Task}
Correlation~\citep{pearl2009causality, pearl2019seven, pearl1995causal} is a statistical measure that describes the size and direction of a relationship between two or more variables. A correlation between variables, however, does not automatically mean that the change in one variable is the cause of the difference in the values of the other variable. Causation \citep{pearl2019seven, pearl2009causality} indicates that one event is the result of the occurrence of the other event; i.e., there is a causal relationship between the two events. This is also referred to as cause and effect.

Theoretically, the difference between the two types of relationships are easy to identify — an action or occurrence can cause another (e.g., having rain droplet causes an increase in the risk of developing rain day), or it can correlate with another (e.g., a visual rain droplet is correlated with a red umbrella, but it does not lead to having a representation of a red umbrella in vision directly). In practice, however, it remains difficult to establish cause and effect, compared with establishing correlation clearly. Yet, most of the current deep learning method focused on directly visualizing the patterns after the visual model is trained without verifying (e.g., intervention methods \citep{pearl2019seven, yang2019causal}) the causation between each representing patterns.

\begin{figure}
  \begin{center}
    \includegraphics[width=0.48\textwidth]{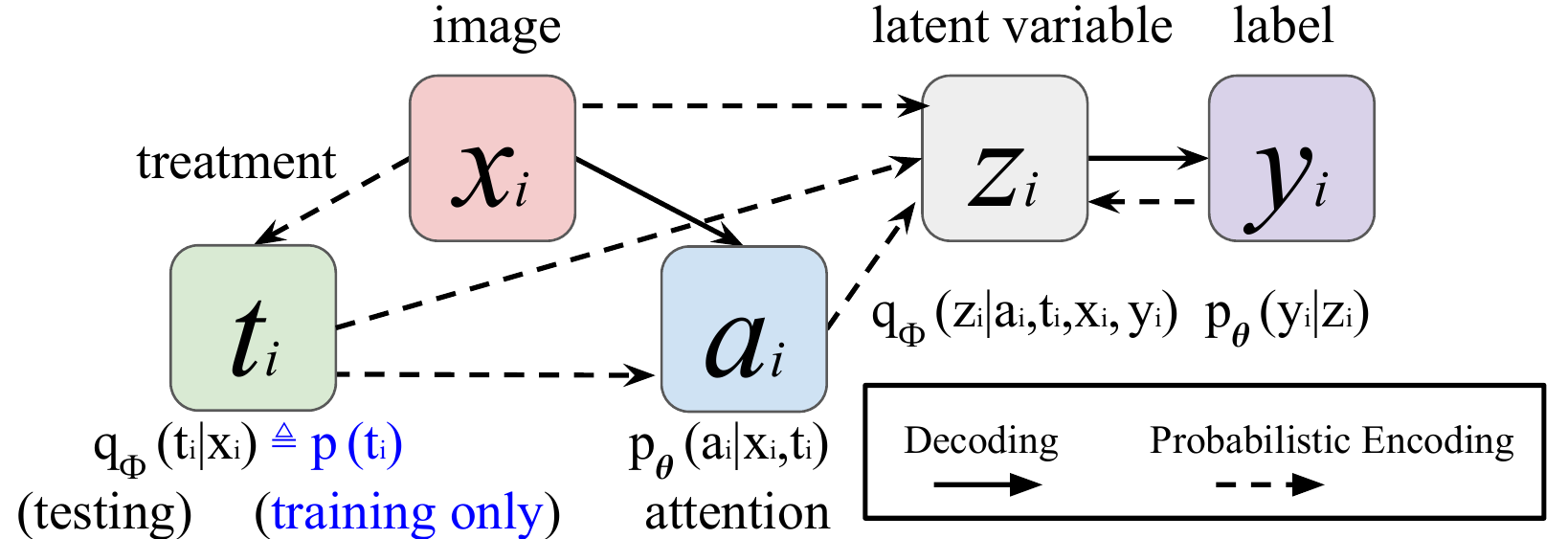}
  \end{center}
  \vspace{-2mm}
  \caption{Generative process of Treatment Inference Transformer model. \textbf{Solid lines}: decoding phase. \textbf{Dashed lines}: probabilistic encoding phase. The treatment information could be only observed during training.}
\label{fig:figure2}
\vspace{-2mm}
\end{figure}

\subsection{Basic Metrics for Causal Inference}
By defining the CE of the individual treatment effect (ITE) as the difference between two potential outcomes for the individual~\citep{louizos2017causal, shalit2017estimating}, the average treatment effect (ATE) is defined as the expected value of the potential outcomes over the subjects.
\section{Dataset}
\label{sup:sect:aba}

\subsection{Dataset Statistics: CPS General}
\begin{figure*}
\centering
\begin{subfigure}{0.5\linewidth}
  \centering
  \includegraphics[width=\linewidth]{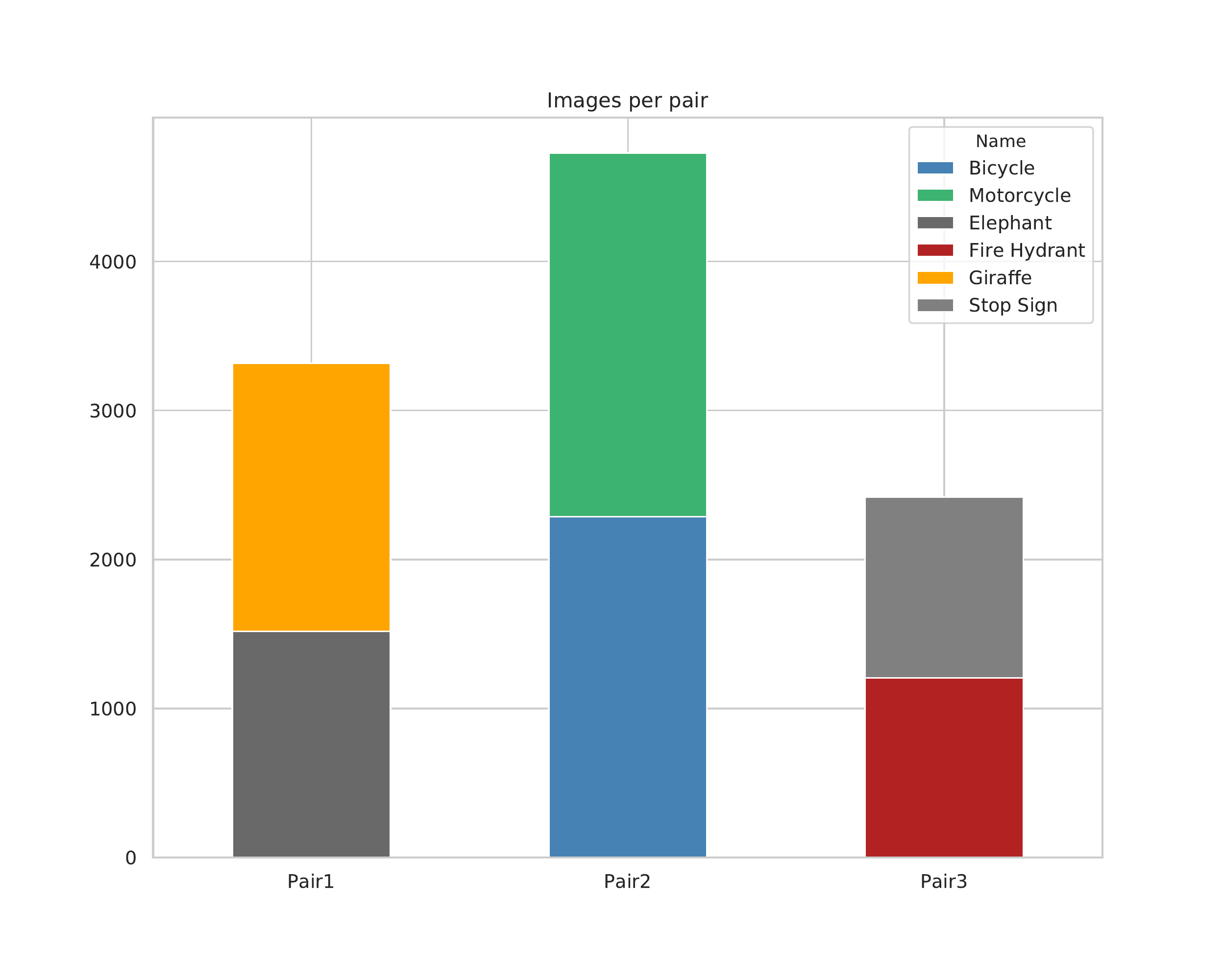}
  \caption{Train Set}
  \label{fig:sub1}
\end{subfigure}%
\begin{subfigure}{.5\linewidth}
  \centering
  \includegraphics[width=\linewidth]{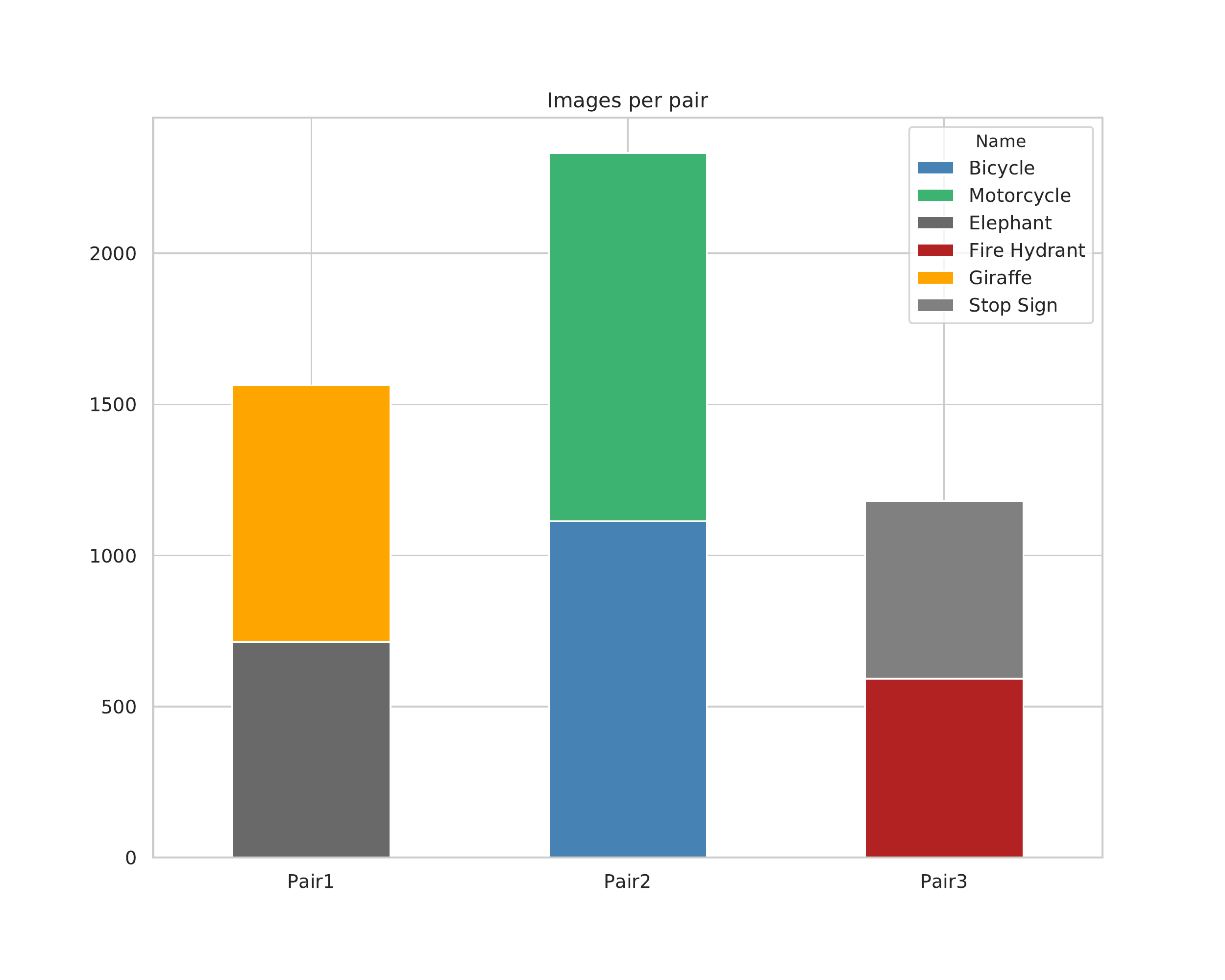}
  \caption{Validation Set}
  \label{fig:sub2}
\end{subfigure}
\caption{{\textbf{Image} details per pairs in CPS: CPS$_1$ -- Giraffe-Elephant, CPS$_2$ -- Bicycle-Motorcycle, and CPS$_3$ -- Stop sign-Fire hydrant.}}
\label{fig:fig2}
\end{figure*}

\begin{figure*}
\centering
\begin{subfigure}{.5\linewidth}
  \centering
  \includegraphics[width=\linewidth]{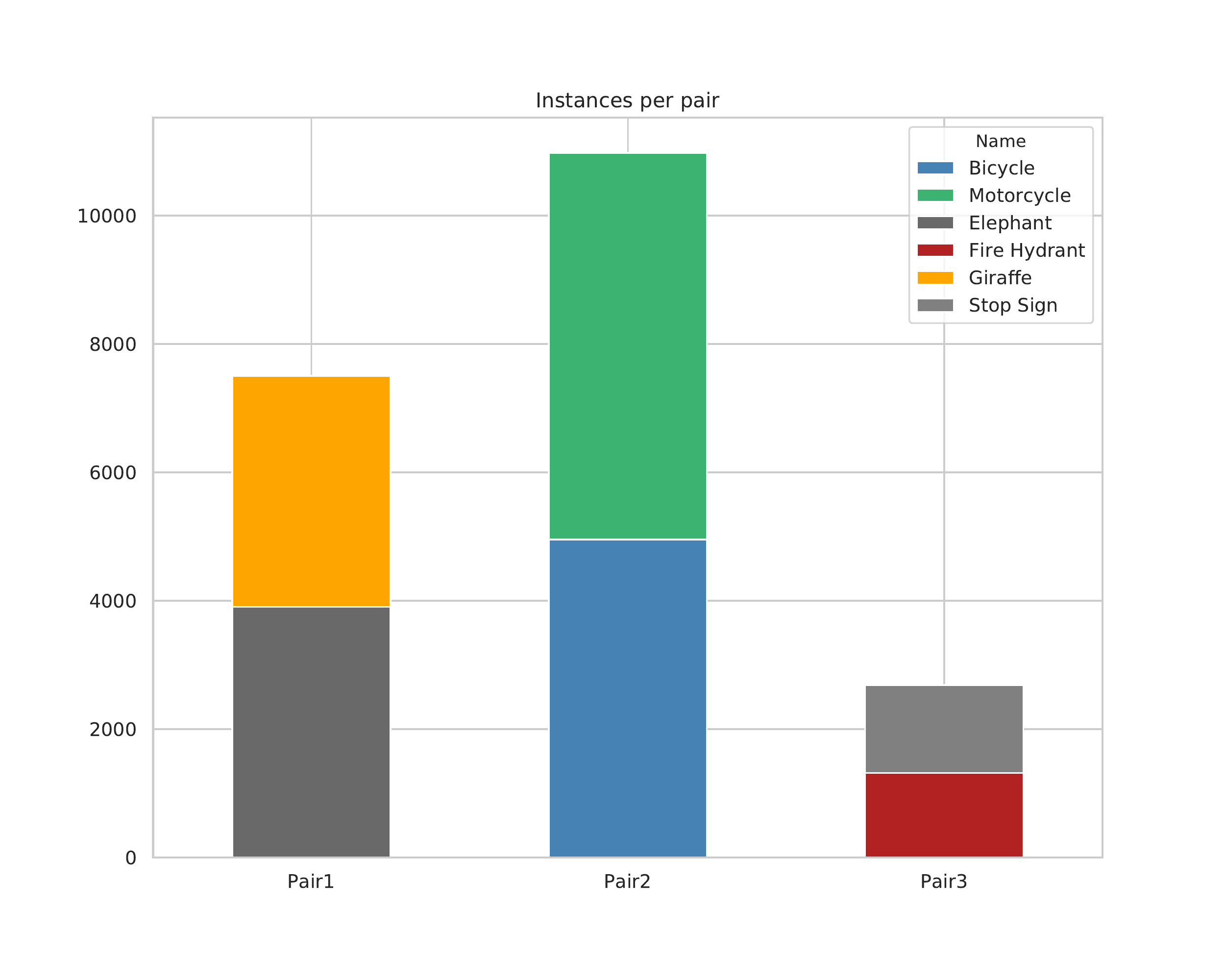}
  \caption{Train Set}
  \label{fig:sub3}
\end{subfigure}%
\begin{subfigure}{.5\linewidth}
  \centering
  \includegraphics[width=\linewidth]{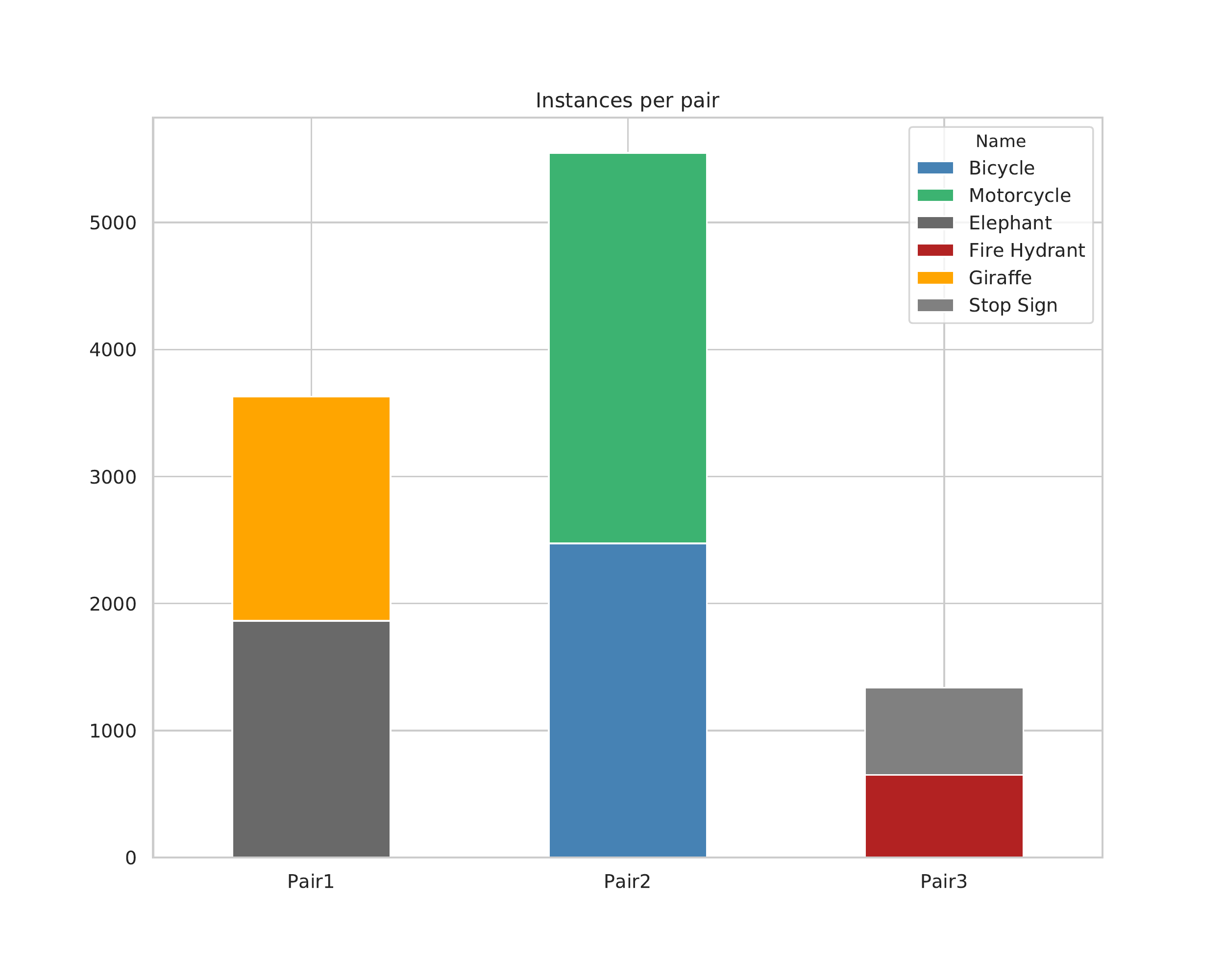}
  \caption{Validation Set}
  \label{fig:sub4}
\end{subfigure}
\caption{\textbf{Instance} details per pairs in CPS: CPS$_1$ -- Giraffe-Elephant, CPS$_2$ -- Bicycle-Motorcycle, and CPS$_3$ -- Stop sign-Fire hydrant.}
\label{fig:fig3}
\end{figure*}

In order to accurately measure visual causality, we select six categories from MS-COCO~\citep{lin2014microsoft} and match them into three causal pairs, which are giraffe-elephant, bicycle-motorcycle, and stop sign-fire hydrant. Each pair is chosen by its matching characteristic and background and split into train and validation set by MS-COCO default setting. With similar characteristics, we are able to study on the causality of how network classifies different categories. To avoid network relying on texture to predict, every pair has a similar background, which makes our classifier more robust to evaluate the causal effect. Next, we analyze the properties of each pair and compare them. The number of images and instances per category for both train and validation set are shown in Fig.\ref{fig:fig2} and Fig.\ref{fig:fig3}. In addition, the segmentation size of each pair is shown in Table.\ref{table:table1}. The segmentation size of each pair influence the performance of our classifier for excessive noise from background increasing the difficulty to find the correct hidden features as shown in Table.\ref{tab:table2}.

\begin{table}[ht!]
\centering
\caption{Percentage of Segmentation in Images}
\begin{tabular}{|l|l|l|}
\hline
\multicolumn{1}{|c|}{\multirow{2}{*}{Pair 1}} & \textbf{Giraffe}   & \textbf{Elephant}     \\ \cline{2-3} 
\multicolumn{1}{|c|}{}                        & 14.60\%            & 24.13\%               \\ \hline
\multirow{2}{*}{Pair 2}                       & \textbf{Bicycle}   & \textbf{Motorcycle}   \\ \cline{2-3} 
                                              & 5.74\%             & 15.59\%               \\ \hline
\multirow{2}{*}{Pair 3}                       & \textbf{Stop Sign} & \textbf{Fire Hydrant} \\ \cline{2-3} 
                                              & 7.58\%             & 7.64\%                \\ \hline
\end{tabular}
\label{table:table1}
\end{table}

\begin{table*}[]
\centering
\caption{\textbf{CPS General:} performance for common object~\citep{lin2014microsoft} causal pairs with different visual treatments.}
\label{tab:table2}
\begin{tabular}{|c|l|l|l|l|}
\hline
\multicolumn{1}{|l|}{}                                                                          & \textbf{Treatment}       & \textbf{CVAE'} & \textbf{CEVAE$_{att}$} & \textbf{TIT}   \\ \hline
\multirow{5}{*}{\textbf{\begin{tabular}[c]{@{}c@{}}Bicycle\\ \&\\ Motorcycle\end{tabular}}}     & Object Masking 0.0       & 78.31 $\pm0.17$        & 80.79 $\pm0.06$          & \textbf{81.05 $\pm0.11$} \\ \cline{2-5} 
                                                                                                & Object Masking 0.5       & 74.98 $\pm0.09$      & 79.46 $\pm0.12$          & \textbf{79.51 $\pm0.16$} \\ \cline{2-5} 
                                                                                                & Object Masking 1.0       & 71.65 $\pm0.23$        & 72.85 $\pm0.13$          & \textbf{73.29 $\pm0.08$} \\ \cline{2-5} 
                                                                                                & Background Refilling 0.5 & 75.28 $\pm0.15$        & 77.5 $\pm0.27$          & \textbf{78.68 $\pm0.20$} \\ \cline{2-5} 
                                                                                                & Background Refilling 1.0 & 71.11 $\pm0.42$        & \textbf{74.49 $\pm0.38$} & 73.95 $\pm0.41$          \\ \hline
\multirow{5}{*}{\textbf{\begin{tabular}[c]{@{}c@{}}Stop Sign\\ \&\\ Fire Hydrant\end{tabular}}} & Object Masking 0.0       & 74.59 $\pm0.29$        & 75.79 $\pm0.26$          & \textbf{77.41 $\pm0.19$} \\ \cline{2-5} 
                                                                                                & Object Masking 0.5       & 72.28 $\pm0.10$        & 73.91$\pm0.05$          & \textbf{74.08 $\pm0.08$} \\ \cline{2-5} 
                                                                                                & Object Masking 1.0       & 68.67 $\pm0.34$        & \textbf{71.22 $\pm0.28$} & 71.06 $\pm0.24$          \\ \cline{2-5} 
                                                                                                & Background Refilling 0.5 & 69.13 $\pm0.16$        & 73.79 $\pm0.21$          & \textbf{75.45 $\pm0.14$} \\ \cline{2-5} 
                                                                                                & Background Refilling 1.0 & 65.62 $\pm0.47$        & 66.65 $\pm0.37$          & \textbf{68.24 $\pm0.44$} \\ \hline
\multirow{5}{*}{\textbf{\begin{tabular}[c]{@{}c@{}}Elephant\\ \&\\ Giraffe\end{tabular}}} & Object Masking 0.0       & 93.72 $\pm0.25$        & 93.53 $\pm0.28$         & \textbf{94.67 $\pm0.20$} \\ \cline{2-5} 
                                                                                                & Object Masking 0.5       & 90.14 $\pm0.11$        & 93.01 $\pm0.19$          & \textbf{93.15 $\pm0.09$} \\ \cline{2-5} 
                                                                                                & Object Masking 1.0       & 80.12 $\pm0.19$        & 82.73 $\pm0.21$ & \textbf{83.06 $\pm0.11$}          \\ \cline{2-5} 
                                                                                                & Background Refilling 0.5 & 90.44 $\pm0.08$        & 91.71 $\pm0.11$          & \textbf{91.73 $\pm0.10$} \\ \cline{2-5} 
                                                                                                & Background Refilling 1.0 & 81.32 $\pm0.28$        & 82.59 $\pm0.29$          & \textbf{83.91 $\pm0.17$} \\ \hline
\end{tabular}
\end{table*}

\subsection{Dataset Statistics: CPS Medical}
In total, 2,633 three-dimensional images (with 658 test images) were collected across multiple anatomies of interest, multiple modalities, and multiple sources (or institutions) representative of real-world clinical applications followed by COCO-CP processing. All images were identified using processes consistent with institutional review board polices at each contributing site. We reformatted the images to reduce the need for specialized software packages for reading to encourage use by specialists in medical imaging for high-level feature reasoning.

\textbf{Human Evaluation} \\
Chest MRI can provide important features to diagnose lung problems such as a tumor or pleural disorder, blood vessel problems, or abnormal lymph nodes. We collaborate with three \textbf{board-certified thoracic surgeons} to review the activate region generated by guided grad-CAM~\citep{selvaraju2017grad} on the test images. The surgeons individually retrospectively reviewed and labeled each study from the generated 100 image results as a DICOM file as consistent or inconsistent saliency compared with their diagnosis using the PACS system. The radiologists have averaged 6.43 years of experience on average, ranging from 5 to 16 years. 

The TIT-generated saliency results also attain the highest consistency (61.2\%) from thoracic surgeons compared with the results from VAE$_{Res}$ (48.1\%) and CEVAE$^{*}_{Res}$ (58.8\%). The consistency from a randomly generated saliency map is only (3.2\%).

\subsection{Ablation Study}
Starting from a ResNet, we modified the architecture towards proposed TIT and compare the accuracy performance of various architecture. Table.~\ref{tab:alba1} shows the impact of each change of the architecture on COCO$_CP$ classification. Among all variation, attention mechanism is the most important feature, while having bilinear fusion (BF) is also more effective than concatenation.

\begin{figure}[ht!]
\begin{center}
   \includegraphics[width=0.99\linewidth]{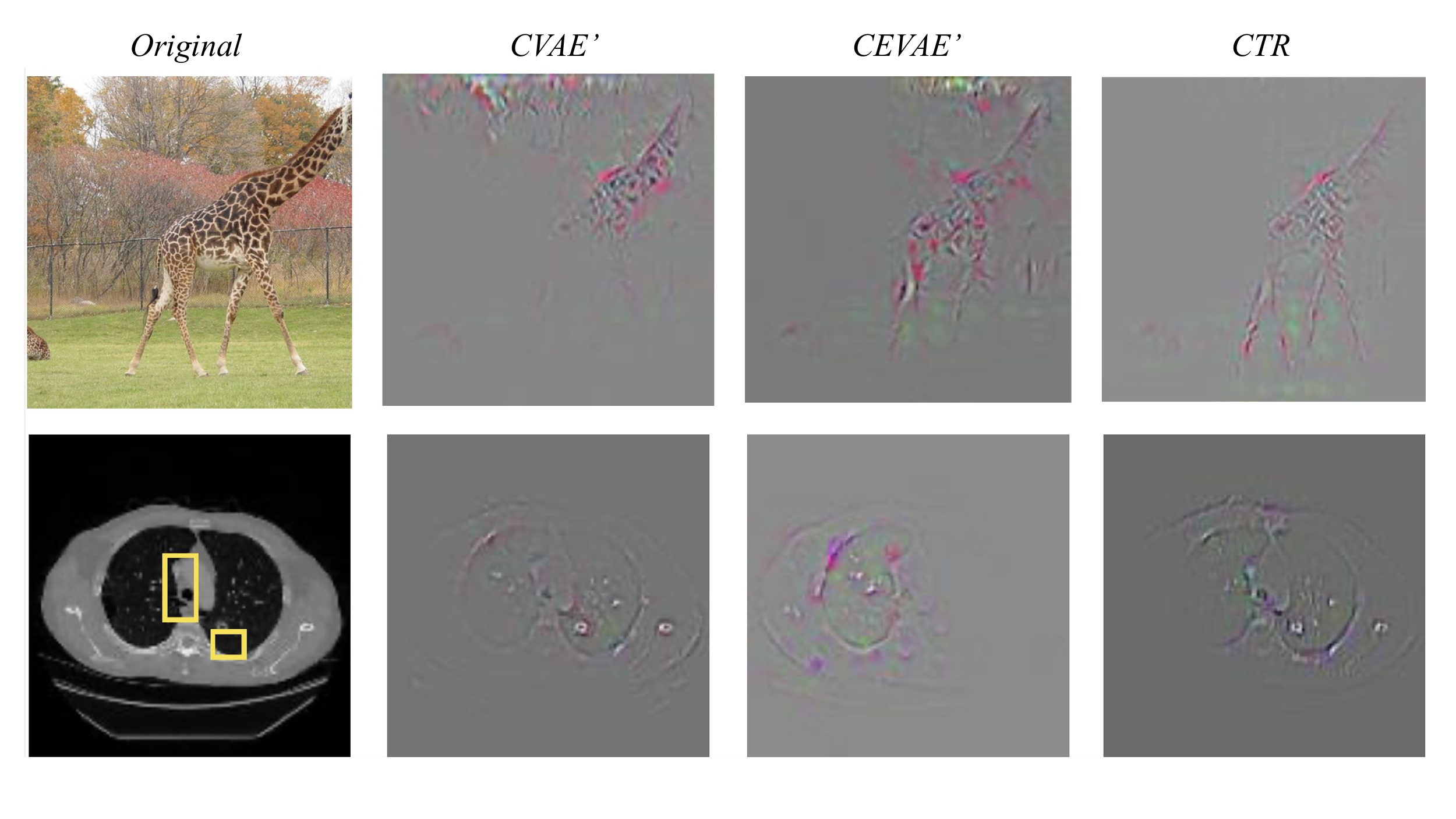}
\end{center}
\vspace{-0.4cm}
\caption{We use class activation mapping methods~\citep{zhou2016learning,selvaraju2017grad} to explain our medical classification model. The yellow bounding box is ground truth label from the Decathlon~\citep{simpson2019large} dataset. The guided-grad CAM method shows a highest false-negative scores on the region of interest. } 
\label{fig:figure:cam1}
\vspace{-0.4cm}
\end{figure}

\begin{figure}[ht!]
\begin{center}
   \includegraphics[width=0.99\linewidth]{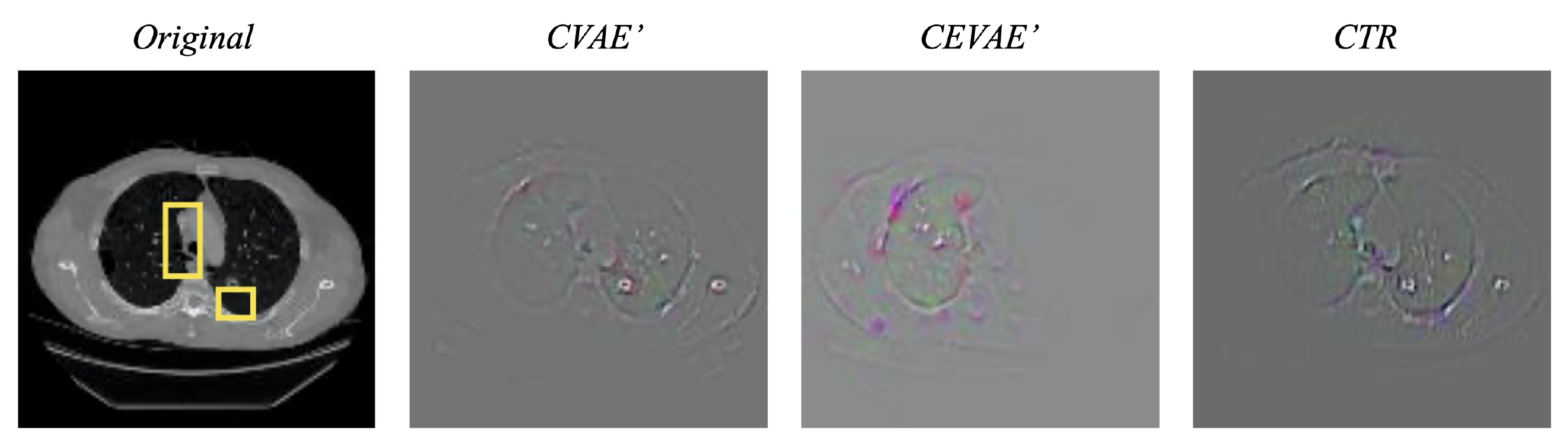}
\end{center}
\vspace{-0.4cm}
\caption{We show Guided-Grad-CAM results on different classification model. The results generated from CAM is much matching yellow bounding box from the Decathlon~\citep{simpson2019large} dataset in the test dataset. } 
\label{fig:figure:gradcam}
\vspace{-0.4cm}
\end{figure}

\begin{table}[h]
\centering
\caption{Accuracy performance under adversarial attack as the treatment. TIT attains higher accuracy in both FGSM, C\&W, and PGD settings. }
\label{tab:adv}
\begin{tabular}{l|lll}
Method & \multicolumn{1}{l|}{CVAE'} & \multicolumn{1}{l|}{CEVAE'} & TIT \\ \hline
FGSM \citep{goodfellow2015laceyella}& 91.92$\pm0.11$  & 92.02$\pm0.11$  & \textbf{92.86$\pm0.12$ } \\
C\&W ~\citep{carlini2017adversarial}& 82.32 $\pm2.34$ & 74.23$\pm4.18$  & \textbf{88.12$\pm1.26$ } \\
PGD ~\citep{madry2017towards}& 74.32 $\pm1.38$ & 86.34$\pm1.70$  & \textbf{89.43$\pm1.08$ }
\end{tabular}
\end{table}

\section{Parameter and Architecture}
\label{sup:arch:sec}

\subsection{Adversarial Perturbation}
With recent security concerns of adversarial example over visual recognition, we also made a broad study on the accuracy and causal effect under adversarial examples. Fast Gradient Sign Method (FGSM) \citep{goodfellow2015laceyella} is a classical gradient-based adversarial noise to generate adversarial  examples by one step gradient update along the direction of the sign of gradient at each pixel by:
\begin{equation}
X_{\text {Adversarial}}=X+\varepsilon \cdot \operatorname{sign}\left(\nabla_{X} J(X, Y)\right),
\end{equation}
where $J$ is the training loss (e.g. cross entropy) and $Y$ is the groundtruth label for $X$.
We adopt FGSM as an visual modification with $\varepsilon$ = $\mathbf{0.3}$ $\ell_{\infty}$ perturbation constraint. This treatment could be further extended on other adversarial examples combined with causal analysis \citep{yang2019causal}. Instead of FGSM, we also study the accuracy performance under Carlini-Wagner attack (C\&W) ~\citep{carlini2017adversarial} and projected gradient descent (PGD) ~\citep{madry2017towards} as treatment. As shown in Table \ref{tab:adv}, our proposed TIT attains higher accuracy and less accuracy degradation in FGSM, C\&W and PGD settings compared to CVAE' and CEVAE' for CPS classification.

\subsection{Overparameterization}
For a fair comparison, we study the performance of architectures with similar number of parameters. To align with the number of parameter in TIT, We modify the number of Resblocks in CVAE' as 4 and add attention mechanism to CEVAE'. As shown in Table \ref{tab:alba2}, with similar number of parameters, our proposed TIT acquire the highest accuracy and better utilize the power of more parameters to compete with the state-of-art CVAE' architecture.

\subsection{Different Mask Size}
We study the effect of different mask sizes with same intervention flipping rate. The object-masking and background-refilling are used as visual perturbation in the experiments. To observe the effect, we gradually increased the mask ratio among the target object. The results in Table \ref{tab:mask:size} and Table \ref{tab:accuracy } show the impact of changing the ratio to the accuracy of CPS general and medical dataset classification. We find the accuracy drops as the ratio increasing, while our proposed TIT is relatively resilient to the high noise ratio scenario and perform better classification.

\subsection{NICO Dataset Settings}
NICO dataset provides several settings to simulate the Non-I.I.D dataset on different levels. 4 typical settings to generate Non-I.I.D training and testing subset.\\
\textbf{Minimum Bias}\\
The setting choose the images in target class as positive samples and images in other classes as negative samples ignoring the context, which could lead to a minimum distribution shift in training and testing subset.\\
\textbf{Proportional Bias}\\
The setting takes all context into consideration but the ratio of each context are different in training and testing subset. In this setting, the level of distribution shift can be adjusted based on the difference of context ratio.\\
\textbf{Compositional Bias}\\
In this setting, the contexts exist in test subset are not guaranteed to exist in training subset. The distribution shift is higher between training and testing set. The shift could be enhanced by adding proportional bias.

We carefully observe the effect of different settings imposing on different model. The result in Table.~\ref{tab:nico:acc:2} shows our proposed TIT performs better in all settings compared to other models including the CNBB model proposed by ~\citep{he2020towards}.

\begin{table}[ht!]
            \centering
            \caption{Model architecture ablation study in COCO$_{CP}$}
            \label{tab:alba1}
            \begin{tabular}{|l|l|}
\hline
\textbf{Architecture} & \textbf{Val. Acc. (\%)} \\ \hline
ResNet & 81.23$\pm0.12$ \\ \hline
ResNet + CVAE = CVAE' & 82.31$\pm0.13$\\ \hline
ResNet + CEVAE = CEVAE'' & 82.17$\pm0.24$ \\ \hline
CEVAE + BF - bernoulli = CEVAE' & 82.68$\pm0.15$\\ \hline
Treatment Inference Transformer (TIT) & \textbf{84.32$\pm0.07$} \\ \hline
\end{tabular}
\end{table}

\begin{table}[ht!]

            \centering
            \caption{Overparameterization ablation study in COCO$_{CP}$}
            \label{tab:alba2}
            \begin{tabular}{|c|l|l|}
\hline
Model & Para. & Val. Acc. (\%) \\ \hline
CVAE' & 4.03M & 82.31$\pm0.13$ \\ \hline
CVAE' + 2 Resblocks & 5.92M & 82.38 $\pm0.18$\\ \hline
CVAE' + 4 Resblocks & \textbf{7.83M} & 81.96 $\pm0.19$\\ \hline
CEVAE' + Attention$_C$ & 7.81M & 83.62 $\pm0.21$\\ \hline
TIT (ours) & 7.39M & \textbf{84.92}$\pm0.07$\\ \hline
\end{tabular}
\end{table}

\begin{figure}[t]
\begin{center}
   \includegraphics[width=0.75\linewidth]{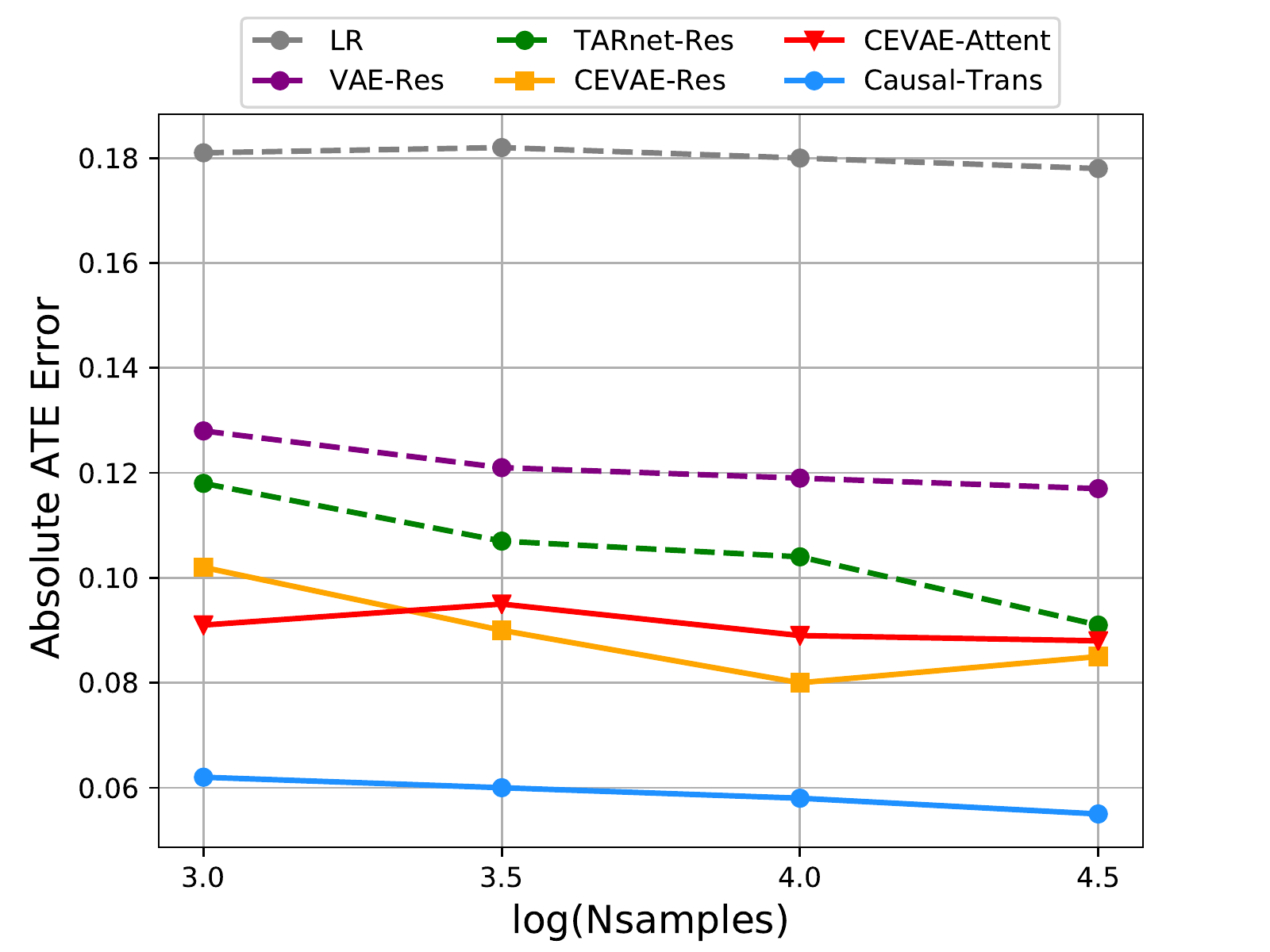}
\end{center}
\caption{Error bar of average treatment effect} 
\label{fig:figure:bar:ate}
\end{figure}

\begin{table}[ht!]
\centering
\caption{Classification accuracy (\%) on NICO dataset with different model and setting. Setting 1,2 and 3 refers to minimum bis, proportional bias and compositional bias mentioned in \ref{sup:arch:sec} separately.}
\label{tab:nico:acc:2}
\begin{tabular}{|c|c|c|c|c|}
\hline
\textbf{Setting} & \textbf{CNBB}~\citep{he2020towards} & \textbf{CVAE'} & \textbf{CEVAE'} & \textbf{TIT} \\ \hline
Setting 1 & 42.96 $\pm1.54$ & 48.72 $\pm2.01$ & 53.94 $\pm1.74$ & \textbf{57.02} $\pm1.42$ \\ \hline
Setting 2 & 44.15 $\pm1.48$ & 50.10 $\pm1.98$ & 54.33 $\pm1.64$ & \textbf{58.75} $\pm1.44$ \\ \hline
Setting 3 & 45.16 $\pm1.52$ & 50.23 $\pm2.12$ & 56.17 $\pm1.82$ & \textbf{60.98} $\pm1.52$ \\ \hline

\end{tabular}
\end{table}

\begin{table*}[ht]
\centering
\caption{Accuracy (\%) of varying mask sizes when intervention flipping rate $\mathbf{n}=0.05$. $\mathbf{I_{OM}}$/$\mathbf{I_{BR}}$ denotes Object-Masking/Background-Refilling. The number $X\%$ means masking $X\%$ of a target object. Our proposed method (TIT) maintains relatively high accuracy as $X$ increases.}
\adjustbox{max width=0.95\textwidth}{
\begin{tabular}{|l|l|l|l|l|l|l|l|}
\hline
\textbf{$\mathbf{I_{OM}}$} & \text{CVAE'} & \text{CEVAE$_{att}$} & \text{TIT} & \text{$\mathbf{I_{BR}}$} & \text{CVAE} & \text{CEVAE$_{att}$} & \text{TIT} \\ \hline
10\% & 93.31 $\pm0.17$& 92.58 $\pm0.21$ & \textbf{94.32$\pm0.08$} & 10\% & 93.04 $\pm0.16$  & 94.25 $\pm0.18$  & \textbf{94.65 $\pm0.09$ } \\ \hline
30\% & 91.19 $\pm0.15$ & 93.37 $\pm0.19$  & \textbf{94.13}$\pm0.07$ & 30\% & 91.27 $\pm0.21$ & 93.53  $\pm0.23$ & \textbf{94.25 $\pm0.11$} \\ \hline
50\% & 90.14 $\pm0.11$ & 93.01  $\pm0.19$ & \textbf{93.15 $\pm0.09$ }  & 50\% & 90.44 $\pm0.08$ & 91.71 $\pm0.11$ & \textbf{91.73 $\pm0.10$ } \\ \hline
70\% & 86.53 $\pm0.13$ & \textbf{91.90 $\pm0.23$ } & 91.26 $\pm0.18$  & 70\% & 86.62 $\pm0.21$ & 88.85 $\pm0.18$ & \textbf{90.46 } $\pm0.12$ \\ \hline
100\% & 80.12 $\pm0.19$ & 82.73 $\pm0.21$  & \textbf{83.06 } $\pm0.11$ & 100\% & 81.32 $\pm0.28$ & 82.59 $\pm0.29$ & \textbf{83.91 }$\pm0.17$ \\ \hline
\end{tabular}
}
\label{tab:mask:size}
\end{table*}
\section{Reproducibility}

\subsection{Hyper-Parameters and Experiment Setup}

    \textbf{C}ausal \textbf{E}ffect \textbf{A}utoencoder\citep{louizos2017causal} \textbf{(CEVAE$^{*}_{Res}$)} baseline: To empower CEVAE for the visual data, our input images use dim(C)=3, dim(X)=128, dim(Y)=128. The encoder part of VAE model utilized in paper takes the ResNet34 as feature extractor. Then we sample the \(q(t|x)\) by Bernoulli distribution, and \(q(y|x,t)\) and \(q(z|x,y,t)\) are sampled by  densely connected hidden layer of 512 neurons. Sequentially, the \(Z\) is generated by reparameterization from \(q|t\). The decoder starts from 3 ResBlocks with 512 width for the dim(Z)=512 to reconstruct the \(p(x|z)\) , we further used 5 upsample blocks with 2 times scaling up and convolution layers with [512,256,128,64,32] width. For the last convolution layer, we use reflection padding with width set as 3. Also, we sample the \(p(t|z)\) and \(p(y|t, z)\) by projecting t and \(\mu_y(t)\) through adaptive pooling and the densely connected hidden layer with 512 width. 
    
    CEVAE with Attention baseline (CEVAE$_{Att}$): For a much fair comparison with proposed Treatment Inference Transformer (TIT), we apply the dual attention module in the encoder part of CEVAE to approximate the \(q(z|x,t,y)\). The dual attention module consists of position and channel attention module.
    \begin{itemize}
        \item Global feature: After 1x1 convolution, the input is scaled by 4 times larger with bilinear interpolation.
        \item Position attention: The module outputs the position attention combined by 2 convolution layers with 64 width to calculate the attention.
        \item Channel attention: The module outputs the channel attention by fusing the channel into spatial information and pass the feature to 2 convolution layers with 64 width.
        \item Combination: The output has the channel of 512, the same width as the input.
    \end{itemize}
The weights of the network are initialized with weights from a model pre-trained on ImageNet. The Adam algorithm with standard parameters and learning rate \textbf{0.001} are utilized for optimization. We use mini-batches of size 128 and pick the models with the highest accuracy. All experiments of our model are implemented in PyTorch using an NVIDIA GeForce GTX 2080 Ti GPU with 12GB memory. The training time for each MS-COCO~\citep{lin2014microsoft} causal pair with different visual treatment takes one hour to two hours on average. The reproducible code of CAN networks and a causal graphical model have been provided in the supplementary and will be open source\footnote{Please follow the readme in the supplementary open-source code for more information}. 

\subsection{Cognitive Response to Attention Mechanism}
Cognitive psychology and neuroimaging~\citep{downing2001testing} studies have found a distinct neural response to the different visual scene, as the visual attention mechanism~\citep{luck2000event}. Attention exercises, Luck et al., ~\citep{luck2000event} have been proved to be enhanced learning capacities by executive control and transferring to cognitive abilities. Since images from different categories vary systematically in their visual properties as well as their semantic category, variation in visual property may influence our cognitive process of visual stimuli. The human brain has the ability to distinguish the visual scenes from different categories when categorical perception is impaired. For example, scrambling and masking are used widely when experimenting with visual pattern sensitivity. Although these perturbation preserved many of their visual characteristics, perception of scene categories was severely impaired, which makes scrambling and masking suitable metrics to compare the visual perception process between neural network and the human brain. These experiments~\citep{luck2000event, leonards2000attention} have been validate on adding attention training for a improved learning performance in human education. 

\subsection{Using Saliency Map to Associate Learned Causal Patterns}

To better understand the learned causal patterns from \textbf{TIT}, we use class activation mapping~\citep{zhou2016learning} (CAM) to study \citep{yang2019causal, pearl2019seven, pearl2009causality, pearl1995causal} the causal patterns. CAM removes all fully-connected layers at the end, and including a tensor product (followed by softmax), which takes as input the global-average-pooled convolutional feature maps, and outputs the probability for each class. To obtain the class-discriminative localization map, Grad-CAM computes the gradient of $\mathbf{yc}$ (score for class $\mathbf{c}$) with respect to feature maps $\mathbf{A}$ and importance weights $\mathbf{\alpha_{k} ^{c}}$ of a convolutional layer. Similar to CAM, Grad-CAM \citep{selvaraju2017grad} heat-map is a weighted combination of feature maps, and followed by a ReLU:
\begin{equation}
L_{\mathrm{Grad} \mathrm{CAM}}^{c}=\operatorname{Re} L U\left(\sum_{k} \alpha_{k}^{c} A^{k}\right)
\end{equation}
In our DNN visualization experiment, we use the state-of-the-are CAM method, guide-GradCAM \citep{selvaraju2017grad} for comparing CVAE', CEVAE', and our TIT. guided-GradCAM fuse guided backpropagation and the Grad-CAM visualizations via a point-wise multiplication.

Interestingly, according to the intervened image after class-activation mapping techniques in Fig. 4 in the main context and Fig. \ref{fig:figure:sp2} in the supplementary, we could find out when the area of interest are much central on the texture and edge effect. To reduce the texture dependent variable, we utilize a neural style transfer on the image set before the intervention.  

\begin{table*}[t]
\centering
\caption{Refuting tests of the causal estimate \citep{pearl2019seven} with causal effect (CE) over different treatment. The validation tests show our method is confident since the common random selection (T$_c$) and Subset test (T$_s$) are closed to the original CE, and all the CE results after replacing treatment with a random (placebo) variable (T$_p$) are close to zero.}
\label{tab:appendix}
\begin{tabular}{|r|l|l|l|l|}
\hline
\multicolumn{1}{|l|}{\textbf{Treatment}} & \textbf{Original ATE} & \textbf{Test-Common (T$_c$)} $\uparrow$ & \textbf{Test-Placebo (T$_p$) $\downarrow$} & \textbf{Test-Subset (T$_s$) $\uparrow$} \\ \hline
IS: TIT & 0.288 & 0.288 & \textbf{0.00479} & 0.288 \\ \cline{2-5} 
CEVAE$_{att}$ & 0.2948 & 0.2941637 & 0.0427 & 0.276 \\ \cline{2-5} 
CVAE' & 0.057 & 0.05673101 & 0.0385 & 0.0583 \\ \hline
AT: TIT & 0.036 & 0.035 & 0.012 & 0.035 \\ \cline{2-5} 
CEVAE$_{att}$  & 0.027 & 0.0274 & \textbf{0.0062} & 0.024 \\ \cline{2-5} 
CVAE' & 0.0247 & 0.0242 & 0.01347 & 0.0156 \\ \hline
SB: TIT & 0.2334 & 0.23385 & 0.0253 & 0.238 \\ \cline{2-5} 
CEVAE$_{att}$  & 0.2417 & 0.2431 & 0.0364 & 0.2353 \\ \cline{2-5} 
CVAE & 0.1853368 & 0.1853 & \textbf{0.01157} & 0.1834 \\ \hline
IM: TIT & 0.1855 & 0.1854 & 0.037 & 0.191 \\ \cline{2-5} 
CEVAE$_{att}$  & 0.22 & 0.22 & \textbf{0.0038} & 0.1736 \\ \cline{2-5} 
CVAE' & 0.222 & 0.22285 & 0.0200707 & 0.1609 \\ \hline
ST: TIT & 0.31763 & 0.317641 & 0.0278 & 0.3351 \\ \cline{2-5} 
CEVAE$_{att}$  & 0.3431 & 0.342221 & 0.0225 & 0.3252 \\ \cline{2-5} 
CVAE' & 0.354412 & 0.354334 & \textbf{0.01127} & 0.3257 \\ \hline
\end{tabular}
\end{table*}

\begin{table*}[t]
\centering
\caption{Classification accuracy (\%) with error bars (with 10-fold cross-validation) comparison between different visual treatments under intervention in the \textbf{CPS} dataset. }
\label{tab:accuracy }
\begin{tabular}{|l|l|l|l|}
\hline
\textbf{Treatment} & \textbf{CVAE'} & \textbf{CEVAE$_{att}$} & \textbf{TIT} \\ \hline
Object Masking 0.0 & 93.61 $\pm 0.15$ & 93.31$\pm 0.11$ & \textbf{94.91}$\pm 0.15$ \\ \hline
Object Masking 0.1 & 93.31 $\pm 0.17$& 93.58$\pm 0.21$ & \textbf{94.32} $\pm 0.08$\\ \hline
Object Masking 0.3 & 91.19 $\pm 0.15$& 93.37$\pm 0.19$ & \textbf{94.13} $\pm 0.07$\\ \hline
Object Masking 0.5 & 90.14 $\pm 0.11$& 93.01 $\pm 0.19$& \textbf{93.15} $\pm 0.09$\\ \hline
Object Masking 0.7 & 86.53 $\pm 0.13$& \textbf{91.90}$\pm 0.23$ & 91.26$\pm 0.18$ \\ \hline
Object Masking 1.0 & 80.12 $\pm 0.19$& 82.73 $\pm 0.21$& \textbf{83.06} $\pm 0.11$\\ \hline
Background Refilling 0.1 & 93.04 $\pm 0.16$& 94.25 $\pm 0.18$& \textbf{94.65} $\pm 0.09$\\ \hline
Background Refilling 0.3 & 91.27 $\pm 0.21$& 93.53 $\pm 0.23$& \textbf{94.25} $\pm 0.11$\\ \hline
Background Refilling 0.5 & 90.44 $\pm 0.08$& 91.71 $\pm 0.11$ & \textbf{91.75}$\pm 0.10$ \\ \hline
Background Refilling 0.7 & 86.62 $\pm 0.21$& 88.85 $\pm 0.18$& \textbf{90.46} $\pm 0.12$\\ \hline
Background Refilling 1.0 & 81.32 $\pm 0.28$& 82.59 $\pm 0.29$& \textbf{83.91} $\pm 0.17$\\ \hline
Image Scrambling & 59.42 $\pm 2.19$& 77.3 $\pm 1.17$& \textbf{78.8} $\pm 0.62$\\ \hline
Style Transfer & 67.73 $\pm 2.19$& 68.12 $\pm 1.21$& \textbf{68.29} $\pm 0.42$\\ \hline
Adversarial Example & 91.92 $\pm 0.11$& 92.02 $\pm 0.11$& \textbf{92.86} $\pm 0.12$ \\ \hline
\end{tabular}
\end{table*}

\begin{figure} [ht]
\begin{center}
   \includegraphics[width=0.99\linewidth]{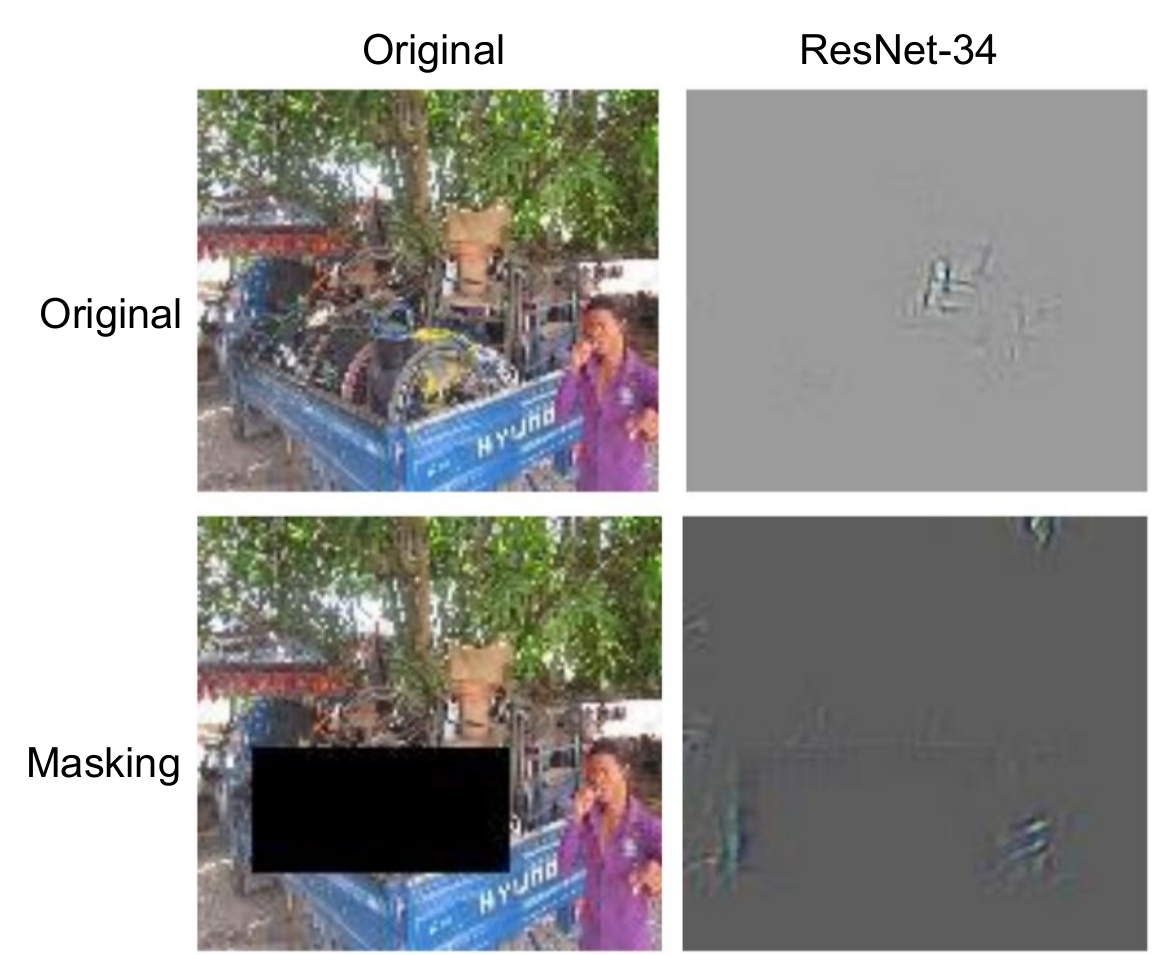}
\end{center}
\vspace{-0.4cm}
\caption{We conduct activation saliency experiment on the COCO-CP trained by bike patterns under partial masking intervention. The guided GradCAM results show current DNNs methods often overfit on the texture and background patterns instead of desired training label(s), which echoes to previous studies~\citep{prakash2018deflecting, yang2019causal, ilyas2019adversarial}.} 
\label{fig:figure:sp2}
\vspace{-0.4cm}
\end{figure}

\section{Identification of Visual Causal Effect}
\label{sup:sec:refu}
\subsection{Causality}
Rubin’s Causal Model (Sekhon, 2008) is a framework developed for the statistical analysis of cause of effect based
on the idea of potential outcomes. Consider:\\
$\bullet ~t_{i},$ a binary treatment $\mathbf{t}$ for individual $i$ with 1 referring to assigning the treatment and 0 to no treatment; \\
$\bullet ~y_{i}$ is the outcome on individual $i$ given a treatment
value. \\
Each individual can have two potential outcomes or (counterfactuals) available as $\{ y_i(1), y_i(0)  \}$  $\mathbf{t}$ corresponding to receiving the treatment or not.  

\textbf{Identifying conceptional treatment effect}
Individual Treatment Effect (ITE) can be defined as the difference
between the two potential outcomes for the individual; Average Treatment Effect (ATE) as the expected
value of the potential outcomes over the subjects. For a binary outcome, it is defined as: given by:
\begin{equation}
y_{i}=y_{i}(0)\left(1-t_{i}\right)+y_{i}(1) t_{i}; \hspace{0.1cm}A T E=\mathbb{E}\left[y_{i}(1)\right]-\mathbb{E}\left[y_{i}(0)\right]; \hspace{0.1cm}
\end{equation}

\begin{flalign}
    y_{i}=y_{i}(0)\left(1-t_{i}\right)+y_{i}(1) t_{i}; \nonumber \\    A T E=\mathbb{E}\left[y_{i}(1)\right]-\mathbb{E}\left[y_{i}(0)\right]; \hspace{0.1cm} 
    \end{flalign}

The above mentioned metric cannot be properly estimated if there are confounding variables in the system, which will introduce bias (Greenland et al., 1999). The causal effect by a treatment variable t on an outcome y is represented by $E[y|do(t = 1)]$, where do represents the fact that the treatment has been kept at a specific value by external interventions on the system which do not affect other variables and their causal relationships in the system. Pearl defines the causal effect for a given treatment t and
an outcome y and other confounding variables $\mathbf{Z}$ as:

Given a proxy X, outcome y, binary treatment t and confounder Z,
we use the back-door criteria to get:
\begin{flalign}
&P(y | X, d o(t=1))= \nonumber \\
&\int_{Z} P(y | X, d o(t=1), Z) P(Z | X, d o(t=1)) d Z
\end{flalign}

Using the intervention manipulation rules, we obtain:
\begin{equation}
    P(y|X, d o(t=1)) 
=\int_{Z} P(y | X, t=1, Z) P(Z | X) d Z.
\end{equation}

The refuting test for all conditional visual model show sustainable performance to the original ATE by random common cause variable test (T$_c$) and random subset test (T$_s$) and an ideally nearby zero ATE results on replacing treatment (T$_r$) with a random variable test. Above validation show our CGM and its associated neural are robust and validated for causal modeling and measurement.

\begin{table}[ht!]
\centering
\caption{Validation of causal effect by three causal refuting tests. The causal effect estimate is tested by random common cause variable test (T$_c$),  replacing treatment with a random (placebo) variable (T$_r$ -- lower is better), and removing a random subset of data (T$_s$). TIT outperforms in most tests.}
\begin{tabular}{|l|llll|}
\hline
Noise : do(t) & \multicolumn{4}{c|}{\textbf{Measurement of ATE }} \\ \hline
\textbf{Method} & \multicolumn{1}{l|}{Original} & \multicolumn{1}{l|}{w/ T$_c$} & \multicolumn{1}{l|}{w/ T$_p$} & w/ T$_s$ \\ \hline
TIT & \textbf{0.2432} & \textbf{0.2431} & \textbf{0.0114} & \textbf{0.2481} \\ \cline{1-1}
CEVAE' & 0.2414 & 0.2414 & 0.0248 & 0.2329 \\ \cline{1-1}
CVAE' & 0.1792 & 0.1763 & 0.0120 & 0.1751 \\ \hline
\end{tabular}
\label{tab:noise}
\end{table}

\section{Evidence Lower Bound of VAE}
To validate an Evidence Lower Bound (ELBO) of our CAN, we assume $\mathbf{p(X, Z)}$, where $\mathbf{X}$ is the observed data and $\mathbf{Z}$ is
the latent representation. $\mathbf{p(X, Z)}$ can be decomposed into
the likelihood and the prior as: $\mathbf{p(X, Z)}$ = $\mathbf{p(X|Z)p(Z)}$.
Using Baye’s inference to calculate the posterior gives:
\begin{equation}
p(Z | X)=\frac{p(X | Z) p(Z)}{\int_{z} p(X | z) p(z)}
\end{equation}
VAE approximates it with the family of distributions $q_{\lambda} (Z|X)$, where $\lambda$ is the variational of parameters for the given family. We minimize the $KL$ divergence to ensure that the approximate distribution used is close to the true posterior:
\begin{equation}
\begin{aligned} K L\left(q_{\lambda}(Z | X) \| p(X | Z)\right)=& \mathbb{E}_{q}\left[\log \left(q_{\lambda}(Z | X)\right)\right]\\ &-\mathbb{E}_{q}[\log p(X, Z)]+\log p(X).\end{aligned}
\end{equation} 
The posterior for inference network will be :
\begin{equation}
q_{\lambda}^{*}(Z | X)=\underset{\lambda}{\arg \min } K L\left(q_{\lambda}(Z | X) \| p(X | Z)\right).
\end{equation}
However, due to the occurrence of $\mathbf{p(X)}$, the KL is still intractable. We can manipulate the above equation by defining the ELBO:
\begin{flalign}
E L B O(\lambda)=\log (p(X))-K L\left(q_{\lambda}(Z | X) \| p(X | Z)\right) \nonumber \\
= \mathbb{E}_{q}[\log p(X | Z)]-K L\left(\log q_{\lambda}(Z | X) \| p(Z)\right).
\end{flalign}

Then, the negative of
the ELBO is the loss function used for the neural networks:
\begin{equation}
\begin{aligned} l(\theta, \phi)=&-\mathbb{E}_{q_{\theta}(z | x)}\left[\log p_{\phi}(X | Z)\right] \\ &+K L\left(\log q_{\theta}(Z | X, \lambda)|| p(Z)\right). \end{aligned}
\end{equation}
 $\theta$ and $\phi $, are the weights and biases of
the DNN which are chosen to maximize the
ELBO using gradient descent algorithm.\\
\textbf{Training Objective of Treatment Inference Transformer.}
In the TIT setting, where the architecture is adapted from TARnet~\citep{shalit2017estimating}’s inference network,
i.e., split input for each treatment group in t after a shared representation, the objective function $\mathcal{L}$ is given by:
\begin{flalign}
\mathcal{L}=\sum_{i=1}^{N} \mathbb{E}_{q\left(\mathbf{z}_{i} | \mathbf{x}_{i}, t_{i}, y_{i}\right)} \left[\log p\left(\mathbf{x}_{i}, t_{i} | \mathbf{z}_{i}\right)+\log p\left(y_{i} | t_{i}, \mathbf{z}_{i}\right)\right] \nonumber\\
+\sum_{i=1}^{N} \mathbb{E}_{q\left(\mathbf{z}_{i} | \mathbf{x}_{i}, t_{i}, y_{i}\right)}[\log p\left(\mathbf{z}_{i}\right)-\log q\left(\mathbf{z}_{i} | \mathbf{x}_{i}, t_{i}, y_{i}\right)]
\end{flalign}
For predicting new subject predictions, the treatment assignment $t$ along with outcome $y$ are required. We have introduced Bernoulli distributions which help predict $y$ and $t$ (a binary index of treatment) for new samples with the theoretical foundation from CEVAE~\citep{louizos2017causal}. We then leverage bilinear fusion for $q(z|x,y,t,a)$ instead of concatenation~\citep{louizos2017causal} and remove Bernoulli sampling for classification label inference. The attention decoding $p(a|x, q(y)) = q(a)$ is incorporating with the known treatment for training. 

\section{Future Work}
\textbf{Discovering Visual Causality beyond Vision Classification Tasks.}\\
In conclusion, we find out causal effect do exist in different DNN-based visual modification methods, and this effect could be visualized to see its effectiveness on understanding targeted DNN layer. 
By introducing a new extended dataset, COCO-CPs, our CAN networks show competitive visualization results and potential combined with existing saliency-based methods. 
For future work, we plan to extend our proposed CAN framework to discover visual causality over more visual tasks, such a video detection, cross-model adaption, and obvious question answering (VQA).

\end{document}